\documentclass{article}
\usepackage{arxiv}

\usepackage[utf8]{inputenc} % allow utf-8 input
\usepackage[T1]{fontenc}    % use 8-bit T1 fonts
\usepackage{hyperref}       % hyperlinks
\usepackage{url}            % simple URL typesetting
\usepackage{booktabs}       % professional-quality tables
\usepackage{amsfonts}       % blackboard math symbols
\usepackage{nicefrac}       % compact symbols for 1/2, etc.
\usepackage{microtype}      % microtypography
\usepackage{lipsum}		% Can be removed after putting your text content
\usepackage{graphicx}
\usepackage{natbib}
\usepackage{doi}

\usepackage{booktabs}
\usepackage{siunitx}
\usepackage{multirow}
\usepackage{graphicx}
\usepackage{enumitem}
\usepackage{bm} % for bold math symbols
\usepackage{tabularx}
\usepackage{subcaption}
\usepackage{float}

\title{Modeling Retinal Ganglion Cells with Neural Differential Equations}
\author{\href{https://orcid.org/0009-0000-2309-1003}{\includegraphics[scale=0.06]{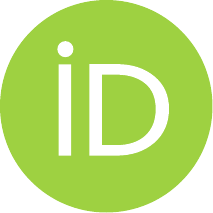}\hspace{1mm}}Kacper Dobek\\\texttt{kdobek@cs.put.poznan.pl} \And \href{https://orcid.org/0009-0007-3070-1632}{\includegraphics[scale=0.06]{orcid.pdf}\hspace{1mm}}Daniel Jankowski\\ \texttt{jankowskidaniel06@gmail.com}  \And \href{https://orcid.org/0000-0001-5439-3231}{\includegraphics[scale=0.06]{orcid.pdf}\hspace{1mm}}Krzysztof Krawiec\\\texttt{krawiec@cs.put.poznan.pl}\thanks{\url{https://ml.cs.put.poznan.pl/en}}\\
Institute of Computing Science, Poznan University of Technology, Poznan, Poland
}

\renewcommand{\headeright}{Technical Report}
\renewcommand{\shorttitle}{Modeling Retinal Ganglion Cells with Neural Differential Equations}
\renewcommand{\undertitle}{Accepted to the AAAI-26 Student Abstract and Poster Program,\\
with supplementary material}

\hypersetup{
pdftitle={Modeling Retinal Ganglion Cells with Neural Differential Equations},
pdfsubject={q-bio.NC, q-bio.QM},
pdfauthor={Kacper Dobek, Daniel Jankowski, Krzysztof Krawiec},
pdfkeywords={Machine Learning, Neural Ordinary Differential Equations, Physics-informed Neural Networks},
}

\begin{document}

\maketitle

\begin{abstract}
This work explores Liquid Time-Constant Networks (LTCs) and Closed-form Continuous-time Networks (CfCs) for modeling retinal ganglion cell activity in tiger salamanders across three datasets. Compared to a convolutional baseline and an LSTM, both architectures achieved lower MAE, faster convergence, smaller model sizes, and favorable query times, though with slightly lower Pearson correlation. Their efficiency and adaptability make them well suited for scenarios with limited data and frequent retraining, such as edge deployments in vision prosthetics.
\end{abstract}

\keywords{Machine Learning, Neural Ordinary Differential Equations, Physics-informed Neural Networks}

\section{Introduction}

\begin{figure}[t]
    \centering
    \includegraphics[width=.9\linewidth]{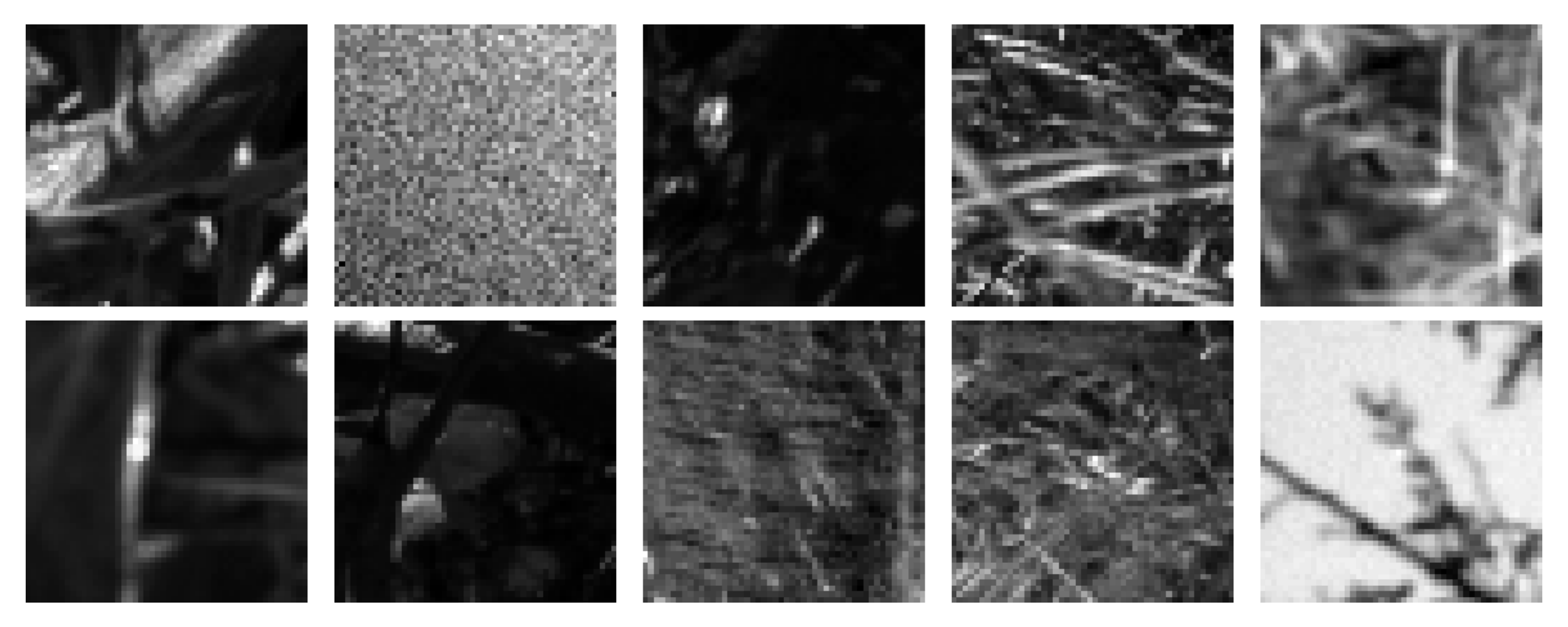}
    \caption{Selected input frames from the dataset.}
    \label{fig:problem-sample-images}
\end{figure}

\begin{figure}[t!]
\centering

\begin{subfigure}[b]{0.42\textwidth}
    \centering
    \includegraphics[width=\textwidth]{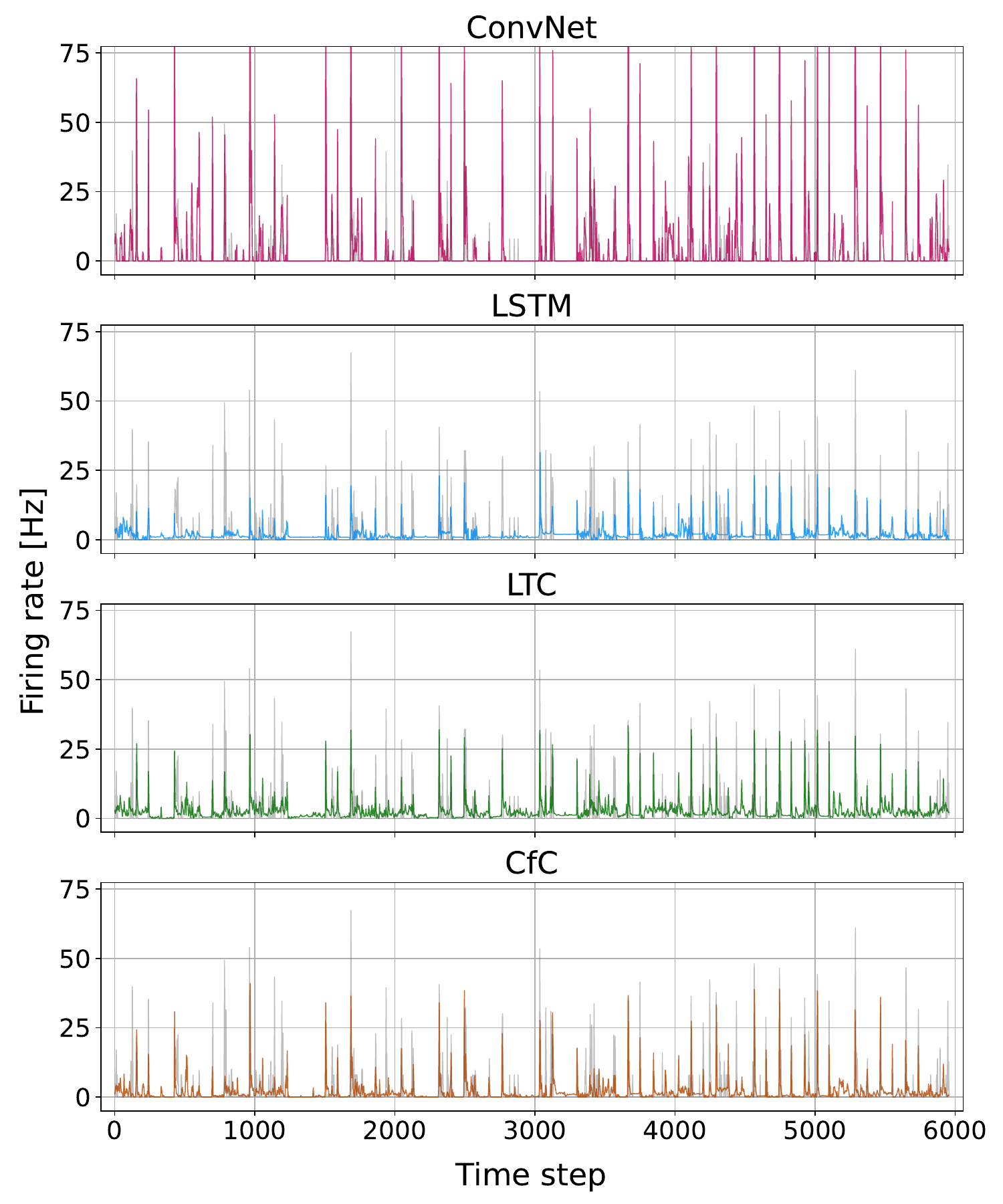}
    \caption{Test-set predictions for the $6$th channel of the rgc9 dataset (ground truth in grayscale).}
    \label{fig:exp-base-reg-comparison}
\end{subfigure}
\hfill
\begin{subfigure}[b]{0.50\textwidth}
    \centering
    \includegraphics[width=\textwidth]{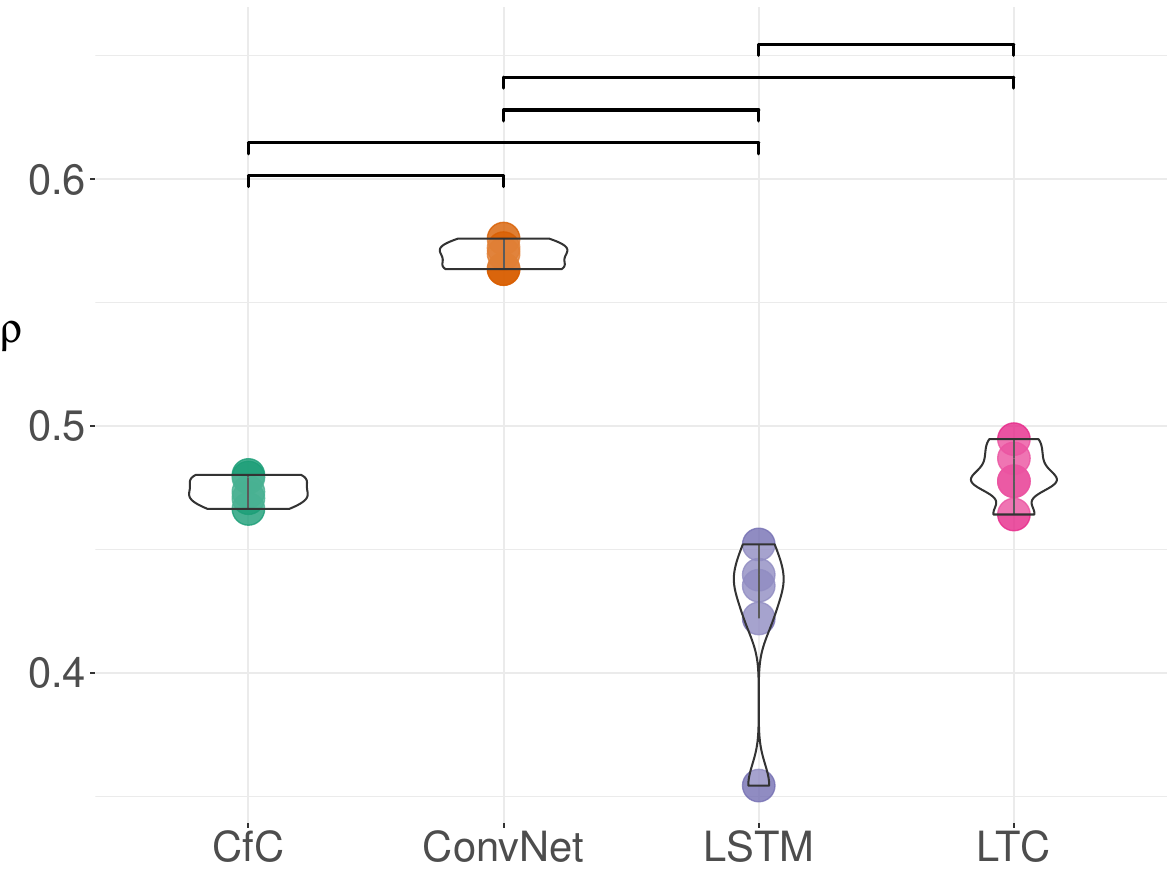}
    \caption{Correlation of the five test-set runs for each model trained on \textit{rgc9} showing significant differences (top)}
    \label{fig:stat-tests}
\end{subfigure}

\caption{Test-set predictions for a selected ganglion cell of \textit{rgc9} (top), and a comparison of test-set correlation between CfC, ConvNet, LSTM, and LTC.}
\end{figure}

Neural Ordinary Differential Equations (NODEs) architectures \cite{theory/chen2019neuralordinarydifferentialequations} promise more precise, fine-grained and plausible modeling of time-dependent phenomena. In this study, we design models based on two recent NODEs, Liquid time-constant neural networks (LTCs) \cite{theory/Hasani20217657} and Closed-form Continuous-time neural networks (CfCs) \cite{theory/Hasani2022992} and apply them to the challenging task of modeling neural activity of retinal ganglion cells (RGCs) of tiger salamanders, following up the work by \citep{retina/maheswaranathan2023interpreting}. Each of the three datasets used therein comprises a monochrome video sequence of natural scenes and synthetic noise images observed by the animal (Fig.\ \ref{fig:problem-sample-images}), accompanied by time series collected from $n=9$, 14, or 27 chosen RGCs (Fig.\ \ref{fig:exp-base-reg-comparison}). Each data point in the time series aggregates neural firings in a 10 ms window; despite this aggregation, the time series are highly irregular and change abruptly (Fig.\ \ref{fig:exp-base-reg-comparison}). The training, validation and test set comprise respectively $280{,}000$, $89{,}802$ and $5{,}996$ frames. See Supplementary Material (SM) Sec.\ \ref{sec:dataset} for more details.

To model RCGs' predictions, Maheswaranathan et al. (\citeyear{retina/maheswaranathan2023interpreting}) designed a simple ConvNet comprising a stack of convolutional layers followed by a dense layer. We replicate this architecture and compare its performance with LTCs, CfCs and LSTMs \cite{theory/Hochreiter19971735}. All architectures are equipped with the same convolution stack processing $40$ consecutive frames concatenated into a single $40$-channel image. For ConvNet, this is followed by a single dense layer, with units corresponding to the $n$ time series. For the remaining architectures, the conv stack is followed by a  32-dimensional dense layer: in LSTM, this latent vector is then fed into an LSTM cell and two dense layers; in NODEs, it becomes direct input to LTC or CfC, configured to match the output dimensionality $n$. The models were trained using Adam optimizer \cite{training/adam} and MSE as the loss function (more details in SM Sec.\ \ref{section:reg-setup}). 
\begin{table}[t!]
\centering
\caption{The correlation coefficient ($\rho$) with $95\%$ confidence intervals and MAE (both aggregated over output channels), the number of parameters, the training time, averaged over $5$ train-test runs (mean $\pm$ std), and the ANOVA $p$-value.}
\begin{tabularx}{\linewidth}{
    l @{\hspace{0.5cm}} l @{\hspace{0.5cm}}
    S[table-format=1.2(2),
    separate-uncertainty = true,
    table-number-alignment = right,
	table-figures-uncertainty = 1,]
    @{\hspace{0.9cm}}
    c
    S[table-format=2.2(2),
    separate-uncertainty = true,
	table-figures-uncertainty = 1,
    table-number-alignment = right]
    S[table-format=6.0,
    table-number-alignment = right]
    S[table-format=4.0(1),
    group-digits = integer,
    group-separator = {,},
    group-minimum-digits = 3,
    separate-uncertainty = true,
	table-figures-uncertainty = 1,
    table-number-alignment = center,
    table-column-width=1.7cm
    ]
    c
}
\toprule
 Dataset & Model & $\rho$\textuparrow & 95\% CI & {MAE}\textdownarrow & {\#params} & {Time [s]} & ANOVA {$p$-value} \\
\midrule
\multirow{4}{*}{\centering\textit{rgc9}} % \rotatebox[origin=c]{90}{}
  & \textit{ConvNet} & 0.569(5) & [0.564, 0.574] & 4.07(90) & 227251 & 2416(196) & \multirow{4}{*}{5.187$\times10^{-8}$} \\
  & LSTM & 0.421(38) & [0.384, 0.457] & 3.18(11) & 46761 & 4759(414) &  \\
  & LTC & 0.480(12) & [0.469, 0.491] & 2.73(12) & 49442 & 1823(360) &  \\
  & CfC & 0.474(06) & [0.468, 0.479] & 2.86(7) & 47496 & 2411(387) &  \\
\midrule
\multirow{4}{*}{\centering\textit{rgc14}}
  & \textit{ConvNet} & 0.586(03) & [0.583, 0.588] & 6.80(264) & 281346 & 2048(326) & \multirow{4}{*}{6.522$\times10^{-7}$} \\
  & LSTM & 0.484(34) & [0.451, 0.517] & 2.72(8) & 49670 & 5234(24) &  \\
  & LTC & 0.569(02) & [0.566, 0.571] & 2.19(4) & 54116 & 1824(159) &  \\
  & CfC & 0.555(09) & [0.546, 0.564] & 2.35(6) & 50752 & 2673(300) &  \\
\midrule
\multirow{4}{*}{\centering\textit{rgc27}}
  & \textit{ConvNet} & 0.586(15) & [0.572, 0.600] & 4.72(189) & 421993 & 2395(262) & \multirow{4}{*}{1.633$\times10^{-12}$} \\
  & LSTM & 0.399(17) & [0.382, 0.416] & 3.62(20) & 53563 & 5019(93) &  \\
  & LTC & 0.529(1) & [0.519, 0.539] & 2.94(6) & 59830 & 2528(474) &  \\
  & CfC & 0.517(07) & [0.511, 0.524] & 3.03(13) & 55144 & 3259(610) &  \\
\bottomrule
\end{tabularx}
\label{tab:regression-results}
\end{table}

\section{Results}

In test-set evaluation (Table \ref{tab:regression-results}), NODEs proved superior to ConvNet and LSTM in terms of MAE on all three datasets. However, they yielded to the ConvNet in terms of correlation coefficient. We tested the normality of residuals and homogeneity of variance, and performed ANOVA for each dataset, revealing a significant difference between the performance of the ConvNet and other models (see Fig.~\ref{fig:stat-tests} for a comparison of \emph{rgc9} results). The reason behind this discrepancy is that NODEs turned out to be better in terms of predicting the \emph{values} of dependent variables, but worse in terms of the \emph{timing} of peaks (Fig.\ \ref{fig:exp-base-reg-comparison}). Notably, NODEs achieve this level of performance with 5x-8x fewer parameters, and require similar training time as the ConvNet, with the LSTM faring worse on this metric.

In the Supplementary Material, we provide the details on the experimental setup (Secs. \ref{sec:dataset}-\ref{sec:exp-node-models}) and report more results. We found NODEs to converge very quickly compared to the other architectures (Sec. \ref{sec:reg-ltc-small-lr}) and have favorable querying times (Sec. \ref{sec:exp-inference-time}). When tested on time series perturbed with noise, ConvNet and LSTM proved more robust (Sec. \ref{sec:reg:noise-robus-exp}). We devised also a variant of the architecture equipped with multiscale temporal representation (Sec. \ref{sec:exp-reg-multiscale-temp}), but it attained worse results than those reported in Table \ref{tab:regression-results}.

\section{Conclusion}

While NODEs considered in this study proved superior to simpler architectures (ConvNet and LSTM) in terms of quantitative prediction (MAE), training time, and model size (and thus interpretability), they did not excel on all metrics, in particular yielding on the time-wise precision of predictions ($\rho$). We hypothesize two main causes behind this phenomenon: (i) the highly irregular and discontinuous nature of considered time series, which may be hard to model with ODEs, and (ii) the apparent absence of complex long-term dependencies in the RGC responses, except for rather obvious and relative constant lag of the neural response (Fig. SM \ref{fig:problem-example-sequence}). Concerning the latter, the constrained, time-agnostic ConvNets might be sufficient to model the essential input-output dependencies present in the data, while being more robust to noise and confounding variables, which form a fair share of the measurement in these observations. NODEs and LSTM, in contrast, are designed to seek more sophisticated temporal dependencies, which may turn out to be spurious and negatively impact generalization in this particular case study. 
As we showed in SM Sec. \ref{sec:reg-base-results}, increasing the input sequence length decreases the performance of NODEs and the LSTM, which may suggest that neuronal responses are characterized by relatively low and and fixed lags, which fit into the adopted 40-frame window and make more intricate modeling unnecessary.
The evidence presented in the SM partially supports these claims, but a more in-depth investigation (possibly involving synthetic data with controlled amount of noise and confounders) is necessary to ascertain this proposition.

An interesting advantage of NODEs demonstrated in this study is their capacity to learn and converge quickly (SM Sec. \ref{sec:reg-ltc-small-lr}). This makes them particularly attractive in scenarios where models need to be quickly or/and frequently updated and where training data is scarce, like edge deployments on specialized hardware, which could be of potential value in vision prosthetics.

\section{Acknowledgments} 

Research supported by the statutory funds of Poznan University of Technology and the Polish Ministry of Science and Higher Education, grant no. 0311/SBAD/0770.

\newpage\appendix
\section*{Supplementary Material}
\section{The dataset} \label{sec:dataset}

\begin{figure}[b]
    \centering
    \includegraphics[width=0.6\linewidth]{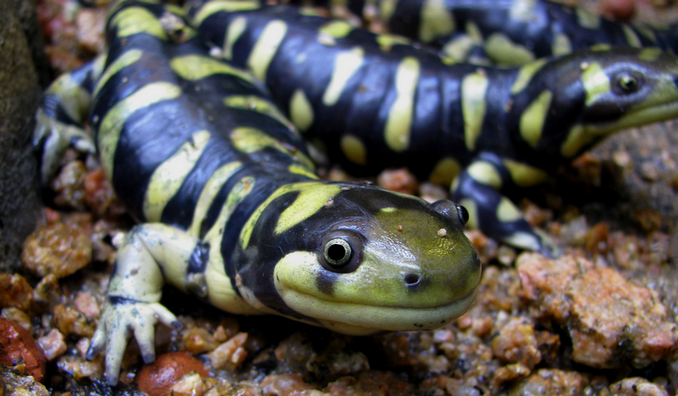}
    \caption{Tiger salamander (\textit{Ambystoma tigrinum}). Image source: \cite{Ribeiro2009}.}
    \label{fig:problem-salamander}
\end{figure}

In this work, we use a publicly available dataset containing recordings of retinal ganglion cell responses of the tiger salamander (\textit{Ambystoma tigrinum}, presented in Fig.~\ref{fig:problem-salamander}) retina to natural and generated visual stimuli, used in the study by Maheswaranathan et al. \citeyear{retina/maheswaranathan2023interpreting}, and available in the Stanford Digital Repository\footnote{\url{https://purl.stanford.edu/rk663dm5577}}.

Responses were obtained from larval tiger salamanders of either sex using a $60$-electrode array for extracellular electrophysiological recordings, as well as sharp microelectrodes for intracellular recordings. Visual stimuli were presented incorporating a video monitor at a refresh rate of $30$ Hz, while the presentation itself was controlled via MATLAB \cite{problem/MATLAB:2010} using the Psychophysics Toolbox~\cite{problem/ThePsychophysicsToolbox}.

The dataset comprises two types of visual stimuli: static natural scenes and generated white noise images. The dataset of natural scenes was generated by converting color images from a natural image database \cite{problem/natural-scene-images} to grayscale.  Pixel intensities were scaled to match the monitor’s minimum and maximum values. These images were then divided into a $50\times50$ grid with each pixel corresponding to a $50 ~\mu m$  region. All regions were randomly selected from the original images without spatial averaging. In each region, a two-dimensional random walk motion was performed with a standard deviation of $0.5$ pixels per frame in horizontal and vertical directions. Moreover, to simulate saccadic movements of the eye, abrupt transitions were applied every second either between different locations within a single frame or between different images; these transitions did not involve smooth sweeping shifts.

As for the second type of visual stimuli, i.e., white-noise images, the dataset includes artificial white noise images. These were binary (black and white) images with the same spatial resolution, duration, and frame rate as the natural stimuli. However, bearing in mind that natural scenes better engage the nonlinear and adaptive computations of the biological visual system, the focus of this work was on natural images.

The authors of  \cite{retina/maheswaranathan2023interpreting} performed three experimental sessions, with $9$, $14$, and $27$ RGCs, respectively. The number of RCGs determines the number of variables to be predicted (also referred to as `channels' in the following). Each session contained responses to both white noise and natural scene stimuli, resulting in six datasets. Since white noise stimuli were not considered in this work, detailed information only about the three datasets used is shown in Table \ref{tab:problem-dataset}.

\begin{table}[b!]
\centering
\caption{Details on the datasets used in this work.}
\label{tab:problem-dataset}
\begin{tabular}{@{}llll@{}}
\toprule
Dataset & Ganglion cells & Number of testing frames & Number of training frames \\ \midrule
\textit{rgc9} & 9 & 5,996 & 359,802 \\
\textit{rgc14} & 14 & 5,996 & 359,802 \\
\textit{rgc27} & 27 & 5,996 & 359,802 \\ \bottomrule
\end{tabular}
\end{table}

Both the stimuli and neural responses were binned in $10$ ms intervals. The dataset was provided with a separate training and test set, which contain $359,802$ ($\sim60$~minutes) and $5,996$ ($\sim5$~minutes) frames, respectively. To train neural networks described further in this work, a validation set containing the last $20\%$ of frames taken from the training set was established manually, resulting in $71,960$ images ($\sim12$~minutes). Sample frames from the dataset are shown in Fig. \ref{fig:problem-sample-images} of the main text.

\begin{figure}[t]
    \centering
    \includegraphics[width=\linewidth, trim={0 0 0 2cm},clip]{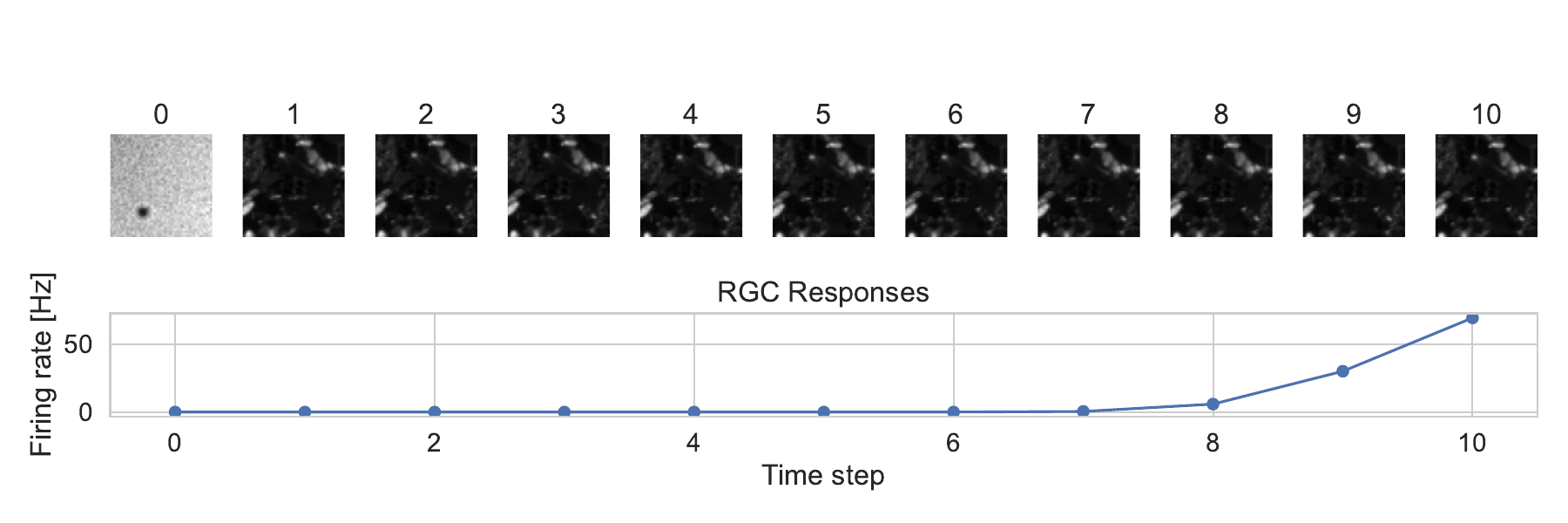}
    \caption{A sample input and output sequence from \textit{rgc9}, channel 1. Observe the frame at time step 0 (left), which has distinct visual characteristics as compared to the following frames. For time step 10, we can see a very high firing rate, suggesting a lag between the change of the stimuli and the RGC signal.}
    \label{fig:problem-example-sequence}
\end{figure}

\begin{figure}[t]
    \centering
    \includegraphics[width=.9\linewidth]{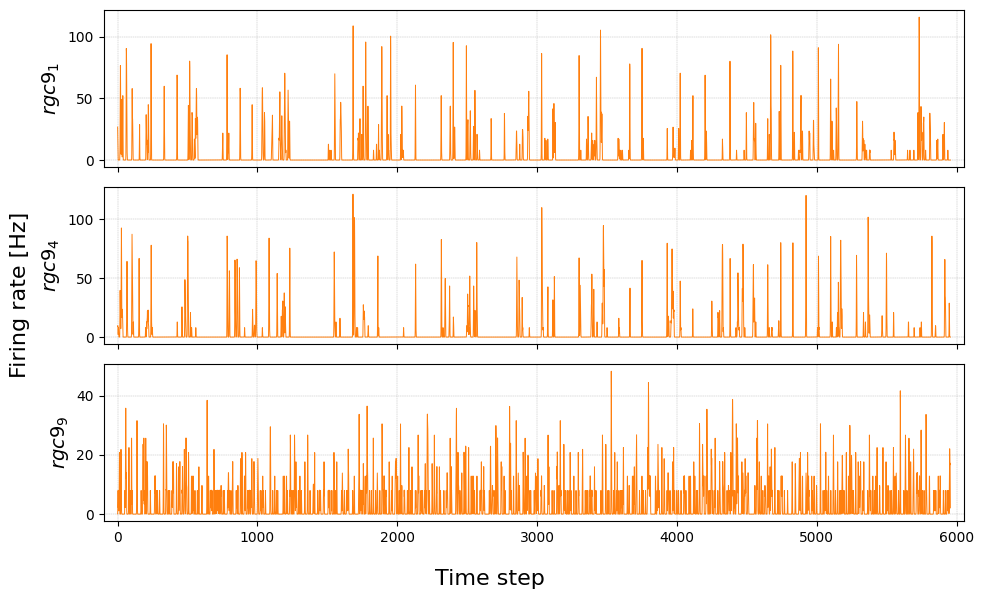}
    \caption{Firing responses to natural stimuli recorded for three selected retinal ganglion cells -- \textit{rgc9\textsubscript{1}}, \textit{rgc9\textsubscript{4}}, and \textit{rgc9\textsubscript{9}} from the test set of \textit{rgc9}. The subscripts refer to the channel numbers (1-based).} 
    \label{fig:problem-test-targets}
\end{figure}

The dataset consists of recordings from a real biological nervous system. As such, it inherently contains biological variability and noise, both from the neural activity itself and the recording process. The firings are sparse, with approximately $80\%$ of the response values  equal to zero. When a high-frequency firing occurs, it is typically brief, with a short onset followed by a rapid return to a low value. Such events occur often shortly after changes in the visual scene, with a characteristic temporal lag -- usually about $8$--$12$ frames later. An exemplary sequence of input images along with the recorded RGC responses is visualized in Fig.~\ref{fig:problem-example-sequence}, while the firings recorded for selected ganglion cells are shown in Fig.~\ref{fig:problem-test-targets}.

\section{Parameters and hyperparameters}
\label{section:reg-setup}

The models reported in the paper were trained and evaluated using the same setting of hyperparameters:

\begin{description}
    \item[Number of epochs = 50:] A sufficient number of iterations over the training set for all considered architectures. Further continuation of learning did not provide an improvement in prediction quality.
    \item[Early Stopping:] The training was terminated after $7$ epochs without improvement of the Pearson correlation coefficient, as assessed on the validation set.    
    \item[Batch size = 4,096:] As the input images were low-resolution grey-scale video frames, it was possible to set a relatively high batch size to ensure efficient memory management and optimal training acceleration on the GPU.
    \item[Prediction step = 1:] Each model was trained to predict the next firing response value based on the history of the previous 40 time steps (frames).
    \item[Dimensionality of hidden state in RNNs:] A hidden state size was the same across all RNN models (LSTM, LTC, CfC) to ensure a fair comparison; however, it was also empirically determined based on the prediction complexity of each dataset, with the number of dimensions set to 16, 24, and 32 for tasks involving 9, 14, and 27 output variables, respectively (Sec.~\ref{sec:dataset}).
    \item[Optimizer:] The Adam optimizer \cite{training/adam} has been used, as it has become a common choice in many deep learning applications.
    \item[Weights Initialization:] Xavier weight initialization \cite{training/xavier} has been used for the designed convolutional encoder to prevent the vanishing gradient problem. For the recurrent neural components, the orthogonal initialization \cite{training/orthogonal} was implemented as it preserves vector norms and improves long-term dependency learning.

    \item[Learning Rate:] Different learning rate values were set for the convolutional encoders and recurrent components. For the encoder block, the learning rate was $0.001$, while for the prediction layer, including the recurrent blocks, it was $0.002$. Additionally, the learning rate was controlled during runtime with a \textit{ReduceLROnPlateau}\footnote{\url{https://docs.pytorch.org/docs/stable/generated/torch.optim.lr_scheduler.ReduceLROnPlateau.html}} scheduler. 
    \item[Target Normalization:] A Min-Max normalization was incorporated separately for each target channel to ensure that the predicted values lie between $0$ and $1$. However, while evaluating the model, the normalized values were rescaled to the original values to compute interpretable and correct metrics.
    \item[Model Checkpointing:] From each experimental run, two states of the model were saved: the best one, which obtained the highest validation Pearson correlation on the validation set, and the final one from the last training epoch. For the evaluation phase, the better of those two was used.
\end{description}

For a given set of parameters, each model was trained from scratch $5$ times, and the results were averaged to provide a reliable estimate of model performance, variability, and robustness. All models were implemented in PyTorch \cite{pytorch} with CUDA support to facilitate more efficient training on Graphical Processing Units (GPUs). Furthermore, our codebase integrates the Weights \& Biases (WandB)~\cite{wandb} AI development platform for tracking and logging of the experiments. To conveniently handle multiple training configurations, we employed Hydra \cite{Yadan2019Hydra}, an open-source Python framework developed by Meta. 

\section{The ConvNet model} \label{sec:reg-convnet}

\begin{figure}[t]
    \centering
    \includegraphics[width=.9\linewidth]{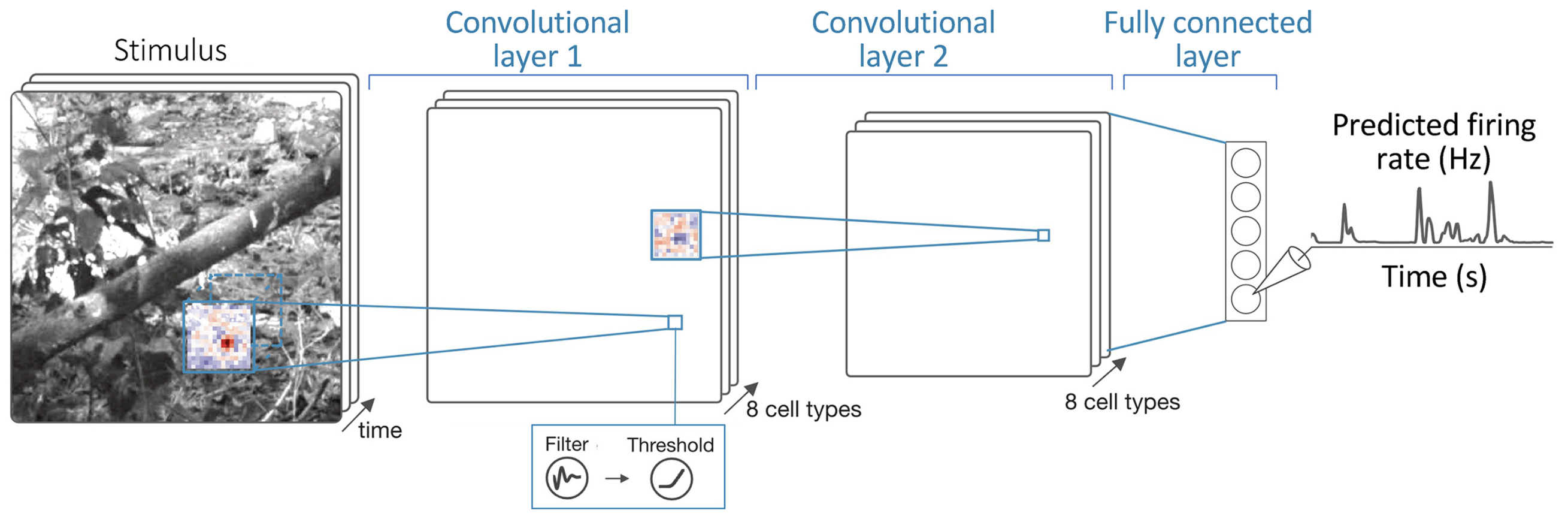}
    \caption{Deep Retina architecture introduced by Maheswaranathan et al. \cite{retina/maheswaranathan2023interpreting} referred to as the \textit{ConvNet}.}
    \label{fig:deep-retina-architecture}
\end{figure}
\begin{figure}[t]
    \centering
    \includegraphics[width=.9\linewidth]{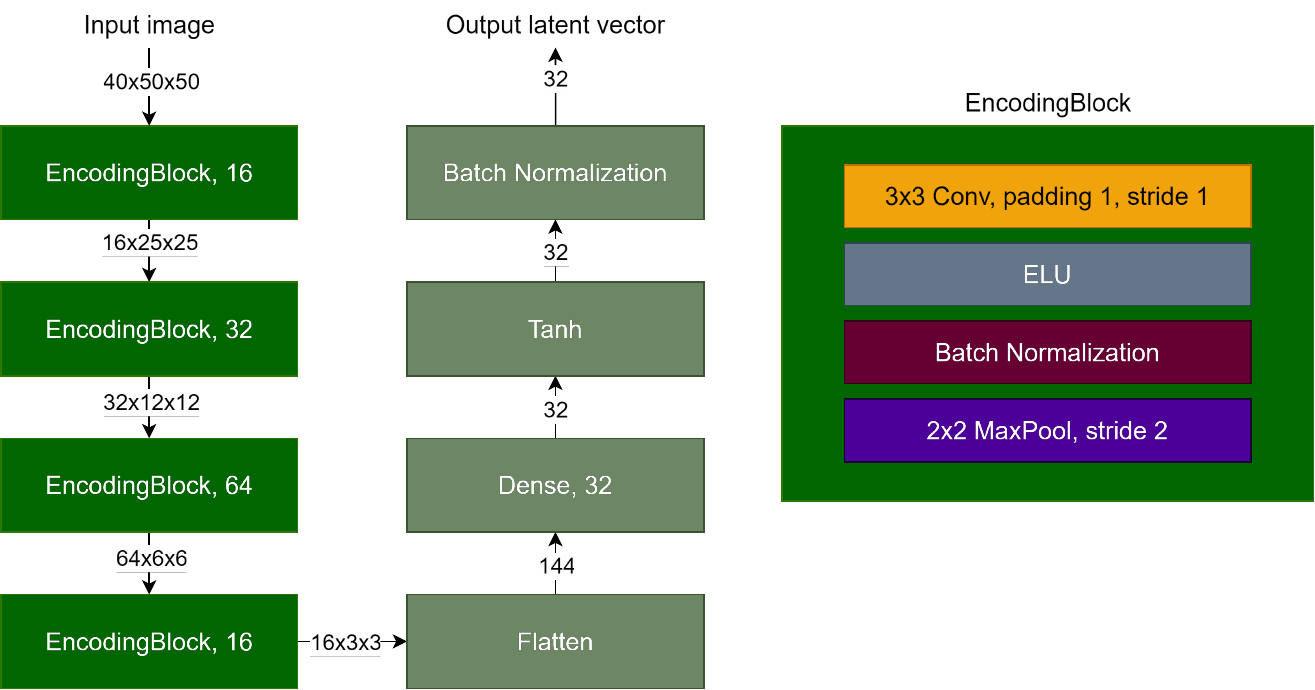}
    \caption{Detailed \textit{encoder} architecture common for all proposed RNNs and Neural ODE-based neural network models. The encoder is composed of four EncodingBlocks.}
    \label{fig:rnns-odes-encoder}
\end{figure}

Designed for modeling retinal ganglion cell responses, the CNN architecture proposed in \cite{retina/maheswaranathan2023interpreting} is a typical \textit{ConvNet} comprising a stack of two convolutional layers followed by a single dense layer (Fig.\ \ref{fig:deep-retina-architecture}). Despite the lack of a recurrent block defined in the architecture, the temporal aspect is encoded directly in the model input. For a given time moment, consecutive $N$ frames were stacked into a single multichannel image. Therefore, the input array/tensor for the model has dimensions $[N, 50, 50]$ ($N$ consecutive grayscale frames of resolution $50 \times 50$ pixels). By default, the $N$ value was set to $40$ by the authors of \cite{retina/maheswaranathan2023interpreting}.

As some details of the proposed \textit{ConvNet} and testing set metric values were not provided in \cite{retina/maheswaranathan2023interpreting}, we attempted to faithfully replicate it to ensure fair comparison to other considered deep neural networks. However, the reconstructed architecture and the training configuration did not achieve the results reported by the authors; therefore, some of its settings have been changed to improve the generalization capabilities. Specifically, the initially adopted Poisson Log-likelihood loss function defined as:
\begin{equation}
    L(y, \hat{y}) = \frac{1}{D}\sum_{i=1}^{D}\hat{y_{i}} - y_{i}\log\hat{y_{i}},
\end{equation}
where
%\begin{eqexpl}[25mm]
$y$ is the measured firing response,
$\hat{y}$ is the predicted firing response,
and $D$ is the batch size,
%\end{eqexpl}
was replaced with Mean Squared Error (MSE).
The adjusted architecture and training configuration allowed the model to achieve metrics very close to those reported in \cite{retina/maheswaranathan2023interpreting}.

\section{LSTM model} \label{sec:reg-lstm}

\begin{figure}[t]
    \centering

    \begin{subfigure}[b]{0.25\linewidth}
        \centering
        \includegraphics[width=\linewidth]{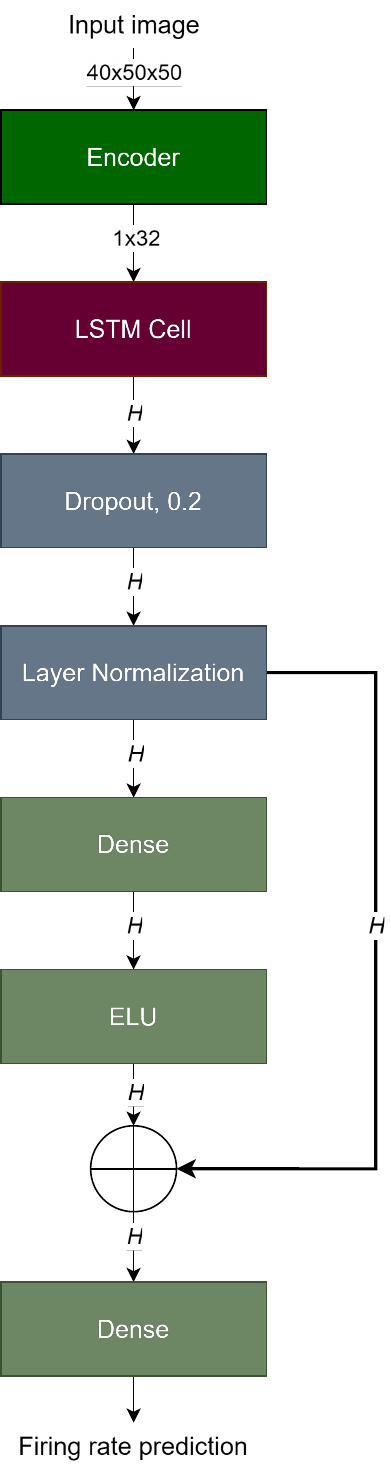}
        \caption{LSTM-based architecture}
        \label{fig:lstm-architecture}
    \end{subfigure}
    \hspace{2cm}
    \begin{subfigure}[b]{0.2\linewidth}
        \centering
        \includegraphics[width=\linewidth]{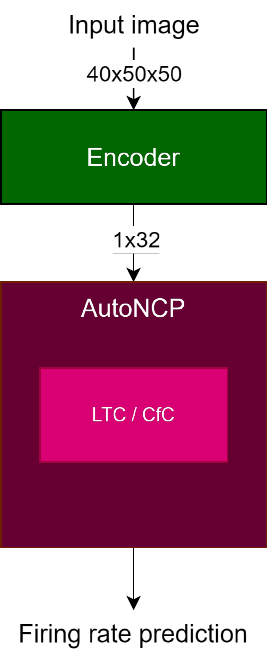}
        \caption{Neural ODE-based architectures: LTC or CfC}
        \label{fig:ltc-cfc-architecture}
    \end{subfigure}

    \caption{Comparison of architecture designs. The encoder block represents the architecture part presented in Fig.\ \ref{fig:rnns-odes-encoder}, while $H$ is the hidden state size, adjusted dynamically based on the dataset complexity as described in Sec.\ \ref{section:reg-setup}.}
    \label{fig:architectures-combined}
\end{figure}

In the default setting, the LSTM model, similarly to the \textit{ConvNet}, starts with processing $40$ consecutive visual frames combined into a single multi-channel image using the conventional convolutional \textit{encoder}. The detailed \textit{encoder} architecture is presented in Fig.\ \ref{fig:rnns-odes-encoder}. The \textit{encoder} was designed from scratch and was common for all considered RNNs and Neural ODE-based models described further. The spatial latent extracted from the \textit{encoder} block was then flattened and served as an input to the LSTM cell, which was followed by two dense layers. (Fig.\ \ref{fig:lstm-architecture}). The architecture incorporated several regularization and stabilization techniques to enhance learning and generalization capabilities. For example, residual connections have been implemented to facilitate gradient flow and mitigate the vanishing gradient problem. 

\begin{table}[t!]
\centering
\caption{All results of the four considered neural network architectures obtained on a test set in the regression task. The notation provided along with the model's name indicates the loss function used and the number of consecutive visual frames combined into a single image. For example, \textit{ConvNet\textsubscript{MAE-20}} indicates that the \textit{ConvNet} model was trained with the MAE loss function, and $20$ historical frames were used as input. The results are averaged over $5$ runs. For each dataset, the model that achieved the highest Pearson correlation coefficient ($\rho$) measured between the model's predictions and targets is marked in bold.}
\label{tab:app:regression-results-all}
\begin{tabular}{
    ll
    S[table-format=1.2(2),
    separate-uncertainty = true,
    table-number-alignment = right,
	table-figures-uncertainty = 1,] 
    S[table-format=2.2(2),
    separate-uncertainty = true,
	table-figures-uncertainty = 1,
    table-number-alignment = right]
    S[table-format=6.0,
    table-number-alignment = right]
    S[table-format=4.0(3),
    separate-uncertainty = true,
	table-figures-uncertainty = 1,
    table-number-alignment = right]
}
\toprule
\textbf{Dataset} & \textbf{Model} & \textbf{$\rho$ \textuparrow} & \textbf{MAE \textdownarrow} & \textbf{\# of params} & \textbf{Time [s]} \\
\midrule
\multirow{16}{*}{\textit{rgc9}}
  & \textbf{\textit{ConvNet\textsubscript{MSE-40}}} & \bfseries 0.57(1) & \bfseries 4.07(90) & \bfseries 227251 & \bfseries 2416(196) \\
  & \textit{ConvNet\textsubscript{MSE-20}} & 0.53(1) & 5.01(202) & 191251 & 1286(238) \\
  & \textit{ConvNet\textsubscript{MAE-40}} & 0.05(4) & 112.00(903) & 227251 & 1288(963) \\
  & \textit{ConvNet\textsubscript{MAE-20}} & 0.04(1) & 102.00(813) & 191251 & 949(779) \\
  & LSTM\textsubscript{MSE-40} & 0.42(4) & 3.18(11) & 46761 & 4759(414) \\
  & LSTM\textsubscript{MSE-20} & 0.35(14) & 3.32(20) & 43881 & 2490(893) \\
  & LSTM\textsubscript{MAE-40} & 0.02(1) & 2.95(39) & 46761 & 1036(120) \\
  & LSTM\textsubscript{MAE-20} & 0.03(1) & 3.54(94) & 43881 & 510(73) \\
  & LTC\textsubscript{MSE-40} & 0.48(1) & 2.73(12) & 49442 & 1823(360) \\
  & LTC\textsubscript{MSE-20} & 0.45(1) & 2.78(2) & 46562 & 1249(65) \\
  & LTC\textsubscript{MAE-40} & 0.06(6) & 2.39(2) & 49442 & 1237(380) \\
  & LTC\textsubscript{MAE-20} & 0.29(1) & 2.28(2) & 46562 & 2128(780) \\
  & CfC\textsubscript{MSE-40} & 0.47(1) & 2.86(7) & 47496 & 2411(387) \\
  & CfC\textsubscript{MSE-20} & 0.44(1) & 2.95(6) & 44616 & 1332(101) \\
  & CfC\textsubscript{MAE-40} & 0.08(10) & 2.40(5) & 47496 & 2246(921) \\
  & CfC\textsubscript{MAE-20} & 0.08(9) & 2.44(9) & 44616 & 1180(757) \\
\midrule
\multirow{16}{*}{\textit{rgc14}}
  & \textbf{\textit{ConvNet\textsubscript{MSE-40}}} & \bfseries 0.59(1) & \bfseries 6.80(264) & \bfseries 281346 & \bfseries 2048(326) \\
  & \textit{ConvNet\textsubscript{MSE-20}} & 0.55(1) & 9.67(360) & 245346 & 1065(174) \\
  & \textit{ConvNet\textsubscript{MAE-40}} & 0.01(4) & 19.04(921) & 281346 & 1015(304) \\
  & \textit{ConvNet\textsubscript{MAE-20}} & 0.01(4) & 16.98(654) & 245346 & 573(89) \\
  & LSTM\textsubscript{MSE-40} & 0.48(3) & 2.72(8) & 49670 & 5234(24) \\
  & LSTM\textsubscript{MSE-20} & 0.41(6) & 2.96(19) & 46790 & 2901(43) \\
  & LSTM\textsubscript{MAE-40} & 0.01(1) & 3.20(38) & 49670 & 968(148) \\
  & LSTM\textsubscript{MAE-20} & 0.02(1) & 3.68(92) & 46790 & 701(286) \\
  & LTC\textsubscript{MSE-40} & 0.57(1) & 2.19(4) & 54116 & 1824(159) \\
  & LTC\textsubscript{MSE-20} & 0.52(1) & 2.46(6) & 51236 & 1151(39) \\
  & LTC\textsubscript{MAE-40} & 0.21(18) & 2.33(13) & 54116 & 3632(982) \\
  & LTC\textsubscript{MAE-20} & 0.07(1) & 2.46(5) & 51236 & 1240(192) \\
  & CfC\textsubscript{MSE-40} & 0.55(1) & 2.35(6) & 50752 & 2673(300) \\
  & CfC\textsubscript{MSE-20} & 0.51(1) & 2.60(15) & 47872 & 1469(290) \\
  & CfC\textsubscript{MAE-40} & 0.03(1) & 2.53(6) & 50752 & 1313(424) \\
  & CfC\textsubscript{MAE-20} & 0.08(10) & 2.47(7) & 47872 & 1245(774) \\
\midrule
\multirow{16}{*}{\textit{rgc27}}
  & \textbf{\textit{ConvNet}\textsubscript{MSE-40}} & \bfseries 0.59(1) & \bfseries 4.72(189) & \bfseries 421993 & \bfseries 2395(262) \\
  & \textit{ConvNet\textsubscript{MSE-20}} & 0.50(1) & 8.90(425) & 385993 & 1198(145) \\
  & \textit{ConvNet\textsubscript{MAE-40}} & 0.04(4) & 51.42(809) & 421993 & 1565(981) \\
  & \textit{ConvNet\textsubscript{MAE-20}} & 0.01(1) & 32.99(932) & 385993 & 987(445) \\
  & LSTM\textsubscript{MSE-40} & 0.40(2) & 3.62(20) & 53563 & 5019(93) \\
  & LSTM\textsubscript{MSE-20} & 0.36(1) & 3.53(6) & 50683 & 2881(18) \\
  & LSTM\textsubscript{MAE-40} & 0.01(1) & 4.70(99) & 53563 & 1172(725) \\
  & LSTM\textsubscript{MAE-20} & 0.02(1) & 4.39(95) & 50683 & 734(329) \\
  & LTC\textsubscript{MSE-40} & 0.53(1) & 2.94(6) & 59830 & 2528(474) \\
  & LTC\textsubscript{MSE-20} & 0.45(1) & 3.22(4) & 56950 & 1304(112) \\
  & LTC\textsubscript{MAE-40} & 0.29(1) & 2.59(4) & 59830 & 4662(785) \\
  & LTC\textsubscript{MAE-20} & 0.18(10) & 2.84(11) & 56950 & 940(32) \\
  & CfC\textsubscript{MSE-40} & 0.52(1) & 3.03(13) & 55144 & 3259(610) \\
  & CfC\textsubscript{MSE-20} & 0.43(1) & 3.38(6) & 52264 & 1430(185) \\
  & CfC\textsubscript{MAE-40} & 0.03(1) & 2.96(4) & 55144 & 1787(925) \\
  & CfC\textsubscript{MAE-20} & 0.28(4) & 2.68(1) & 52264 & 2650(427) \\
\bottomrule
\end{tabular}
\end{table}

\section{Neural ODE-based models} \label{sec:exp-node-models}

The NODE models LTC and CfC comprise an identical encoder (Fig.\ \ref{fig:rnns-odes-encoder}) as the LSTM, but use either an LTC module or a CfC module as the recurrent component. Moreover, the models are integrated into an NCP wiring, a biologically-inspired framework designed by Lechner et al. \cite{theory/ncp} and implemented in the \texttt{ncps} Python package.

The LTC and CfC architectures for modeling retinal ganglion cells accept as input a flattened representation of stacked frames processed by the encoder. In the default scenario, no activation function is used on the final output, as early experimentation with an activation function such as ReLU right after the NCP showcased problems with gradient flow. Nevertheless, ReLU activation was applied during inference, as the firing rate is nonnegative. We present a high-level overview of the architecture in Fig.\ \ref{fig:ltc-cfc-architecture}.

To the best of our knowledge, there exist no studies that would give an intuition on tuning the CfC parameters. As the default values provided by the authors yielded solid results in our experiments, it was decided to use them in the following experimental iterations.

\begin{figure}[t]
    \centering
    \includegraphics[width=.9\linewidth]{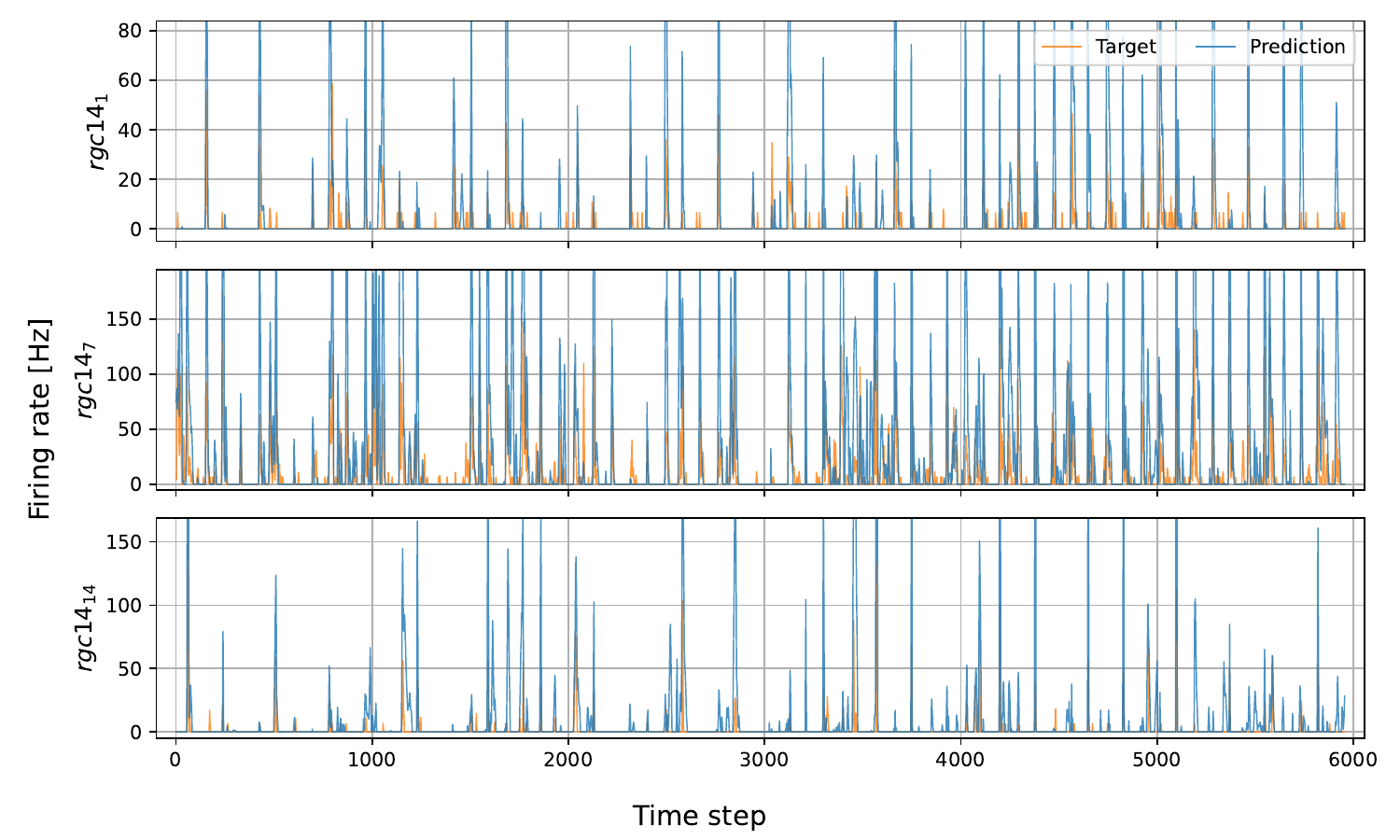}
    \caption{Predictions of the \textit{ConvNet} model obtained from the test set on the $1st$, $7th$, and $14th$ channel of the \textit{rgc14} dataset. Overall, the model achieved a reasonable correlation as shown in Table \ref{tab:regression-results} of the main text; however, it struggled with predicting the correct firing value, reaching the highest MAE among all considered architectures. In this figure, the y-axis for each channel is scaled respectively to the target firing responses. This perspective shows the \textit{ConvNet} model often overestimates ground-truth values.}
    \label{fig:exp-base-reg-convnet-14}
\end{figure}
\begin{figure}[t]
    \centering
    \includegraphics[width=.9\linewidth]{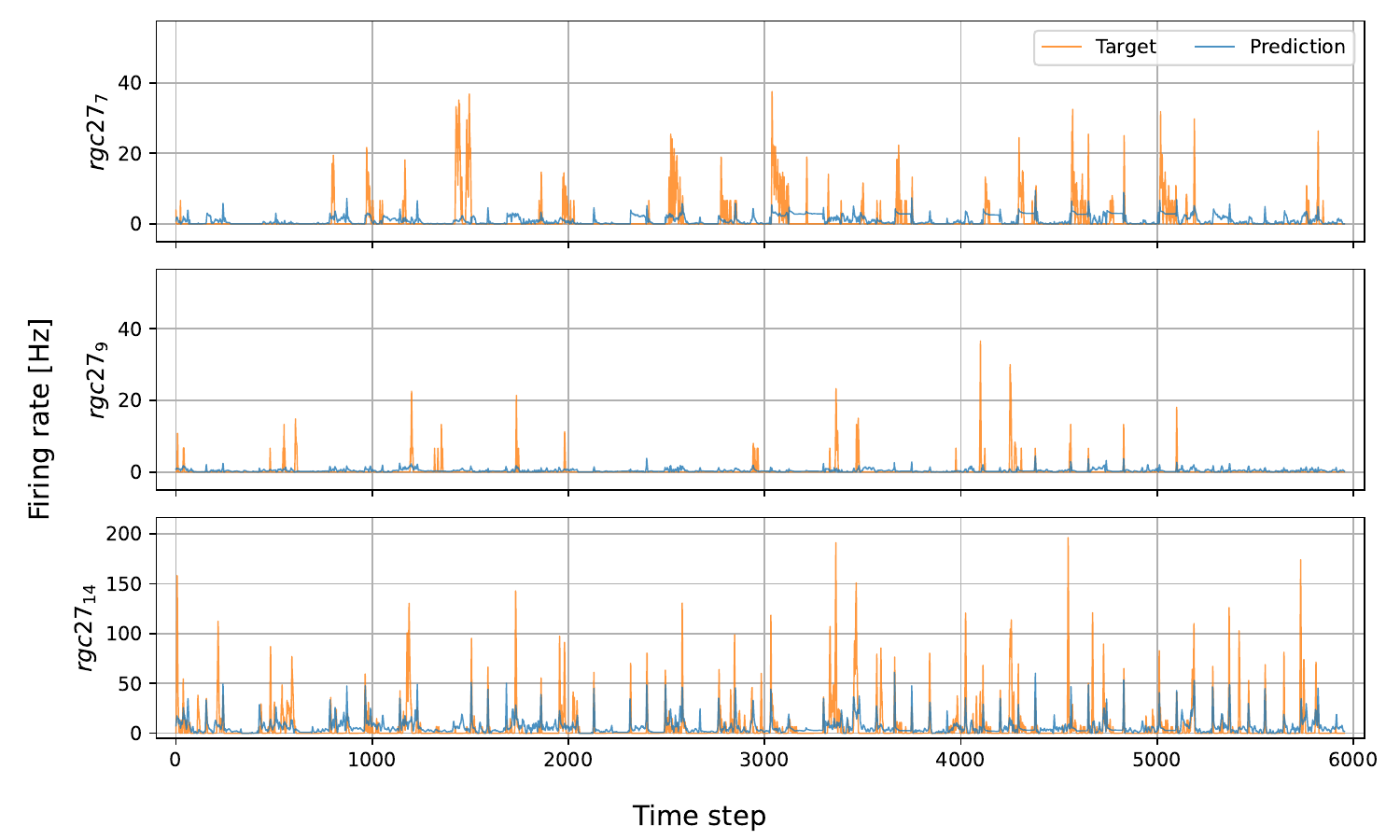}
    \caption{Predictions of the LSTM model obtained from the test set on the $7th$, $9th$, and $14th$ channel of the \textit{rgc27} dataset. The model generalized quite well for cells characterized by high firing rates that are distinctive and relatively frequent (\textit{rgc27\textsubscript{14}}) but had significant problems with cells that fired seldom or where firings are grouped (\textit{rgc27\textsubscript{7}}, \textit{rgc27\textsubscript{9}}).}
    \label{fig:exp-base-reg-lstm}
\end{figure}

\section{Detailed results} \label{sec:reg-base-results}

\begin{figure}[t]
    \centering
\includegraphics[width=.9\linewidth]{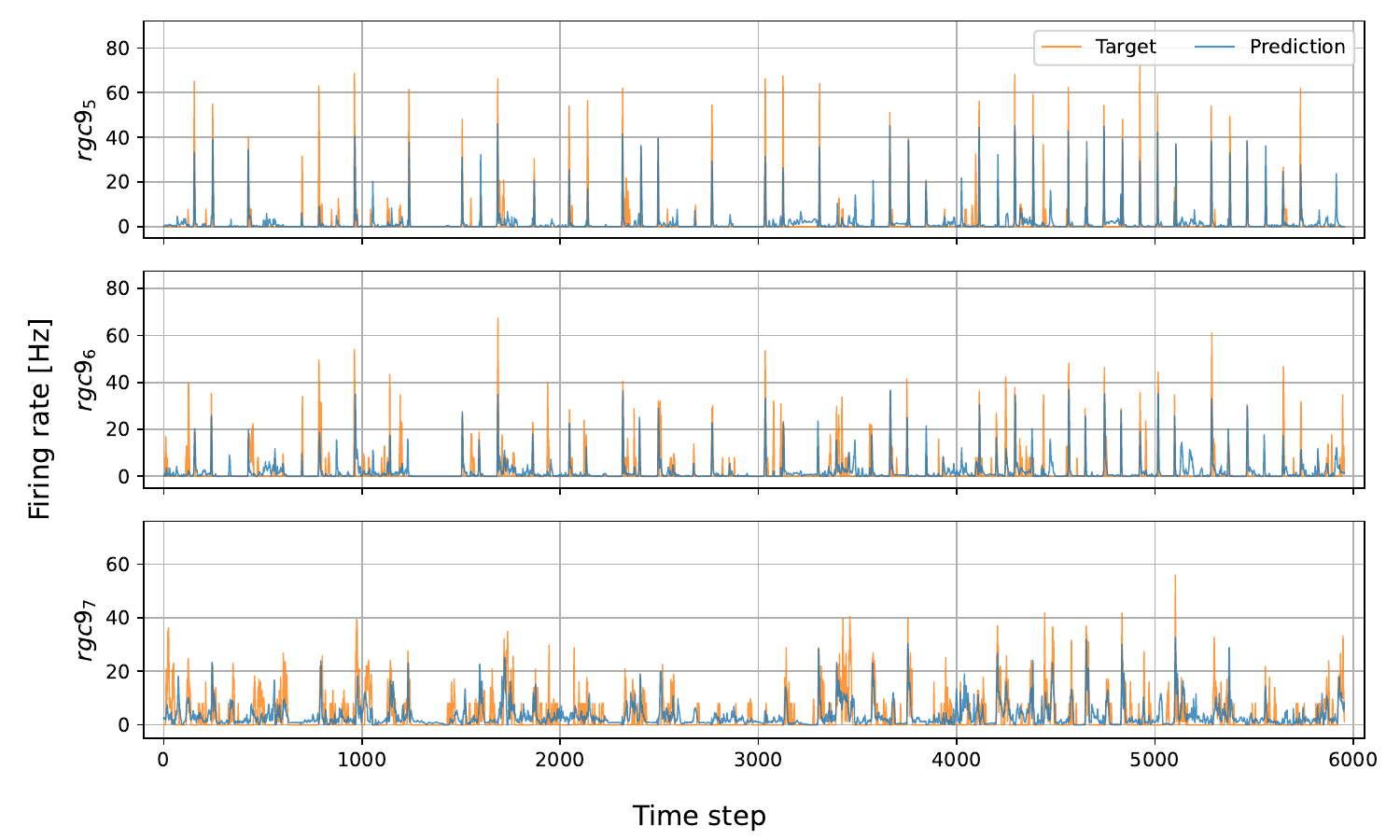}
    \caption{Predictions of the LTC model obtained from the test set on the $5th$, $6th$, and $7th$ channel of the \textit{rgc9} dataset.}
    \label{fig:exp-base-reg-ltc}
\end{figure}
\begin{figure}[t]
    \centering
\includegraphics[width=.9\linewidth]{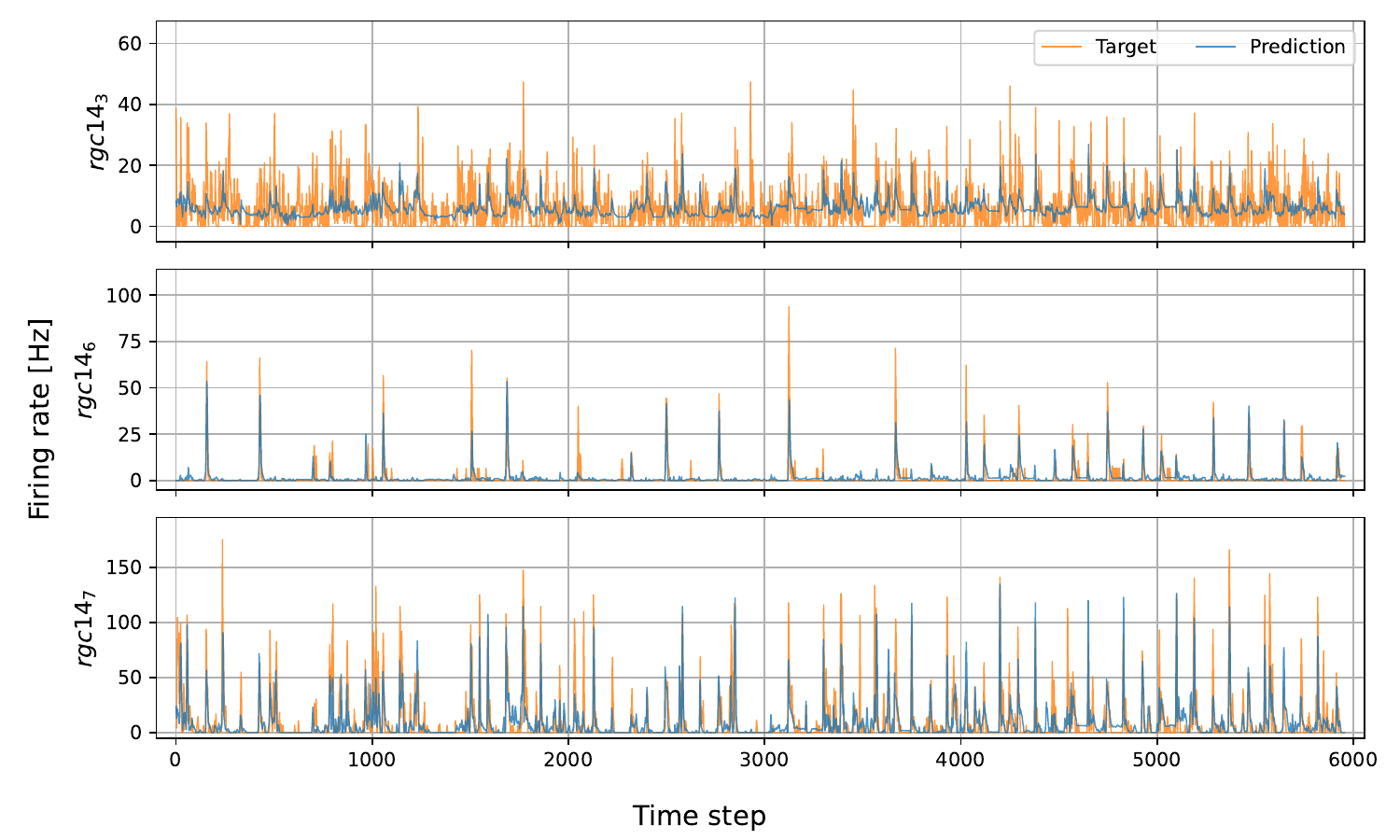}
    \caption{Predictions of the CfC model obtained from the test set on the $3rd$, $6th$, and $7th$ channel of the \textit{rgc14} dataset. There seems to be a constant offset between 0 and the lowest predictions for more `dense' target series (i.e., in the presence of more non-negative values), such as presented in the first row, \textit{rgc14\textsubscript{3}} channel.}
    \label{fig:exp-base-reg-cfc}
\end{figure}

\begin{figure}[t]
    \centering

    \begin{subfigure}[b]{0.32\textwidth}
        \centering
        \includegraphics[width=\textwidth]{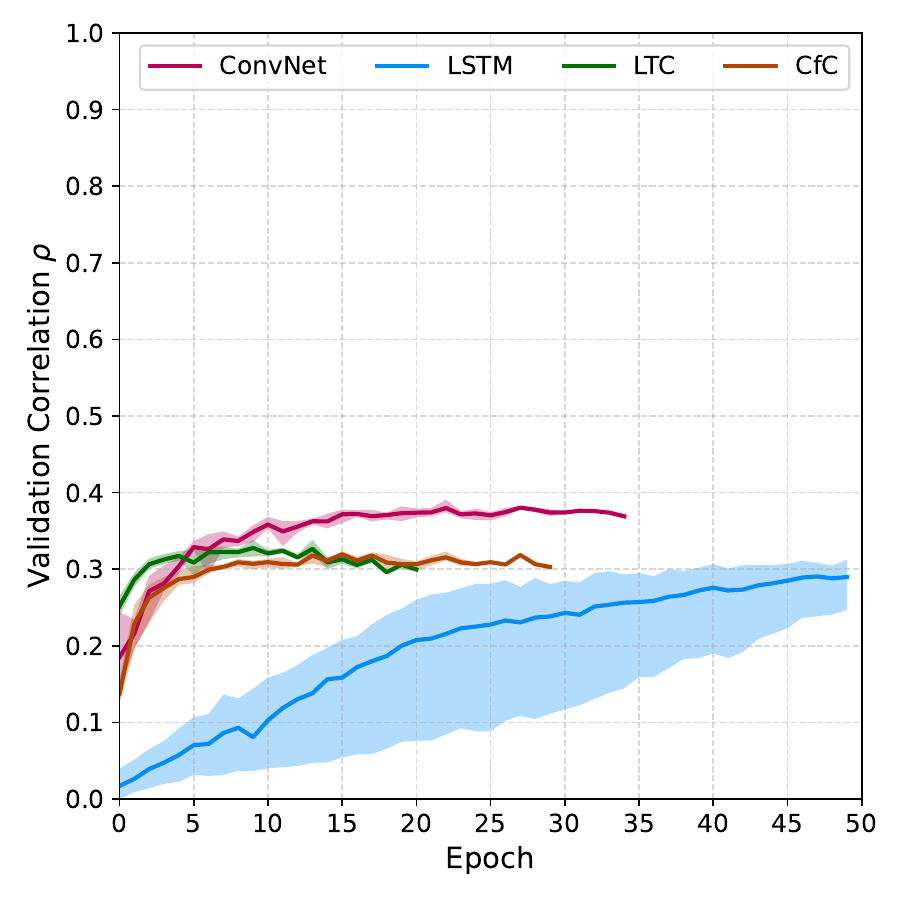}
        \caption{\textit{rgc9}}
        \label{fig:exp-reg-valid-cor-9}
    \end{subfigure}
    \begin{subfigure}[b]{0.32\textwidth}
        \centering
        \includegraphics[width=\textwidth]{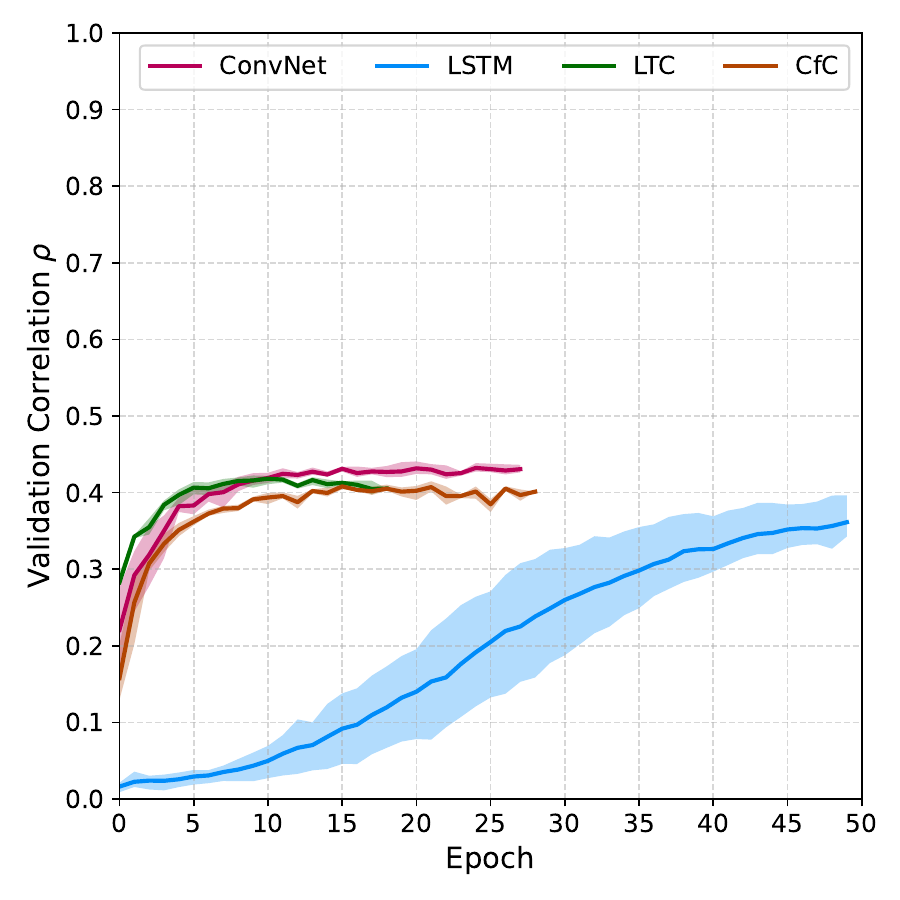}
        \caption{\textit{rgc14}}
        \label{fig:exp-reg-valid-cor-14}
    \end{subfigure}
    \begin{subfigure}[b]{0.32\textwidth}
        \centering
        \includegraphics[width=\textwidth]{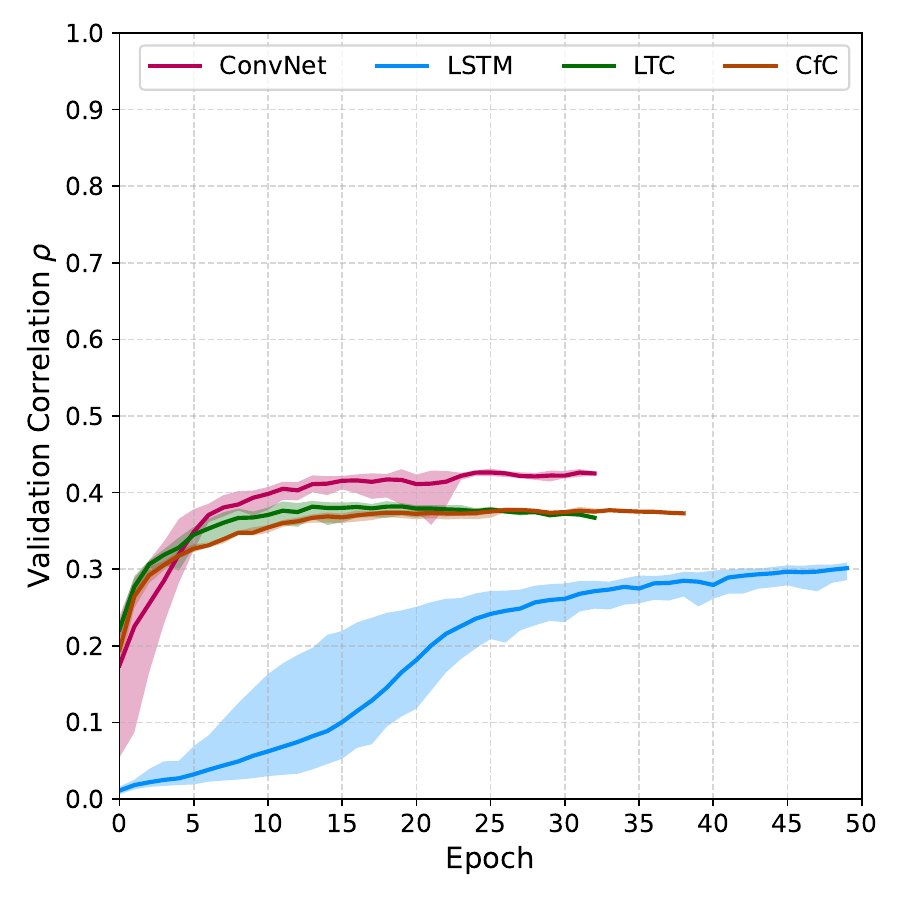}
        \caption{\textit{rgc27}}
        \label{fig:exp-reg-valid-cor-27}
    \end{subfigure}

    \caption{Averaged correlation curves obtained on validation sets by models during the training process.}
    \label{fig:exp-reg-valid-cor}
\end{figure}

To maintain clarity of the presented results, Table \ref{tab:regression-results} in the main text shows only the aggregated outcomes for the best configuration of each model. In Table\ \ref{tab:app:regression-results-all}, we demonstrate the detailed results, including the setups with different loss functions and different numbers of visual frames at input. The results indicate that the MAE loss function is not effective in training models on this task. Moreover, as expected, reducing the number of historical frames provided to the model results in worse model performance.

The \textit{ConvNet's} predictions for selected ganglion cells, collected from a test set of \textit{rgc14} dataset, are presented in Fig.\ \ref{fig:exp-base-reg-convnet-14}. Remarkably, the highest correlation obtained by this model was not reflected in MAE, where the recurrent architectures achieved better results, especially the LTC model, which turned out to be the best in this metric across all datasets. The main reason for this phenomenon is that the \textit{ConvNet} model, although it correctly predicts the firing time, often assigns it a value that is significantly higher than the target. Figure~\ref{fig:exp-base-reg-ltc} presents the LTC model predictions for selected channels of the \textit{rgc9} dataset. Proceeding to the results across all evaluated models, Fig. 2 of the main text shows the predictions of the $6th$ ganglion cell from the test set of the \textit{rgc9} dataset. The observation that RNN architectures achieve lower MAE values compared to the \textit{ConvNet} model is also reflected in the visual analysis for exemplary cells. Even though the predictions of these models are not correlated as much as in the case of the \textit{ConvNet} model, the predicted values themselves better align with the ground-truth firing responses. Additional visualizations presenting the individual predictions of LSTM and CfC models are shown in Figs.~\ref{fig:exp-base-reg-lstm} and \ref{fig:exp-base-reg-cfc}, respectively. Remarkably, the LSTM model achieved promising modeling accuracy for some cells, especially those that fire singly and relatively frequently; however, it struggled with most channels, where firings occurred suddenly and had lower magnitude.

\begin{figure}[t]
    \centering

    \begin{subfigure}[b]{0.45\textwidth}
        \centering
        \includegraphics[width=\textwidth]{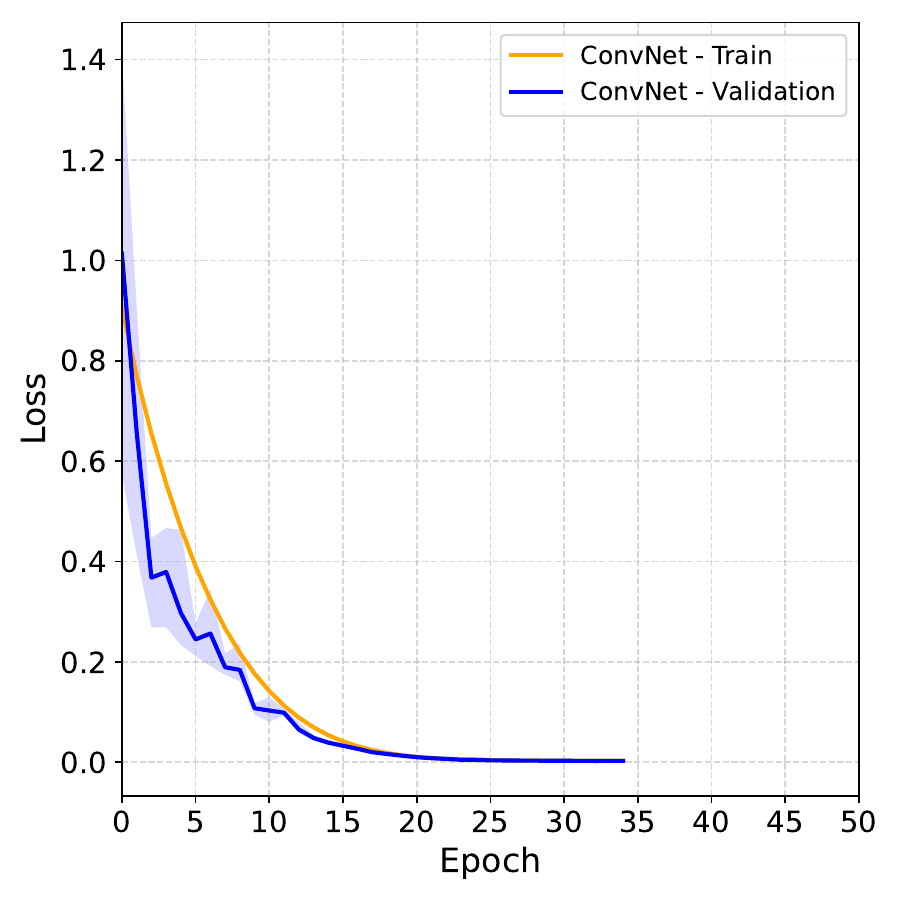}
        \caption{\textit{ConvNet}}
        \label{fig:exp-reg-convnet-curve}
    \end{subfigure}
    \hspace{5mm}
    \begin{subfigure}[b]{0.45\textwidth}
        \centering
        \includegraphics[width=\textwidth]{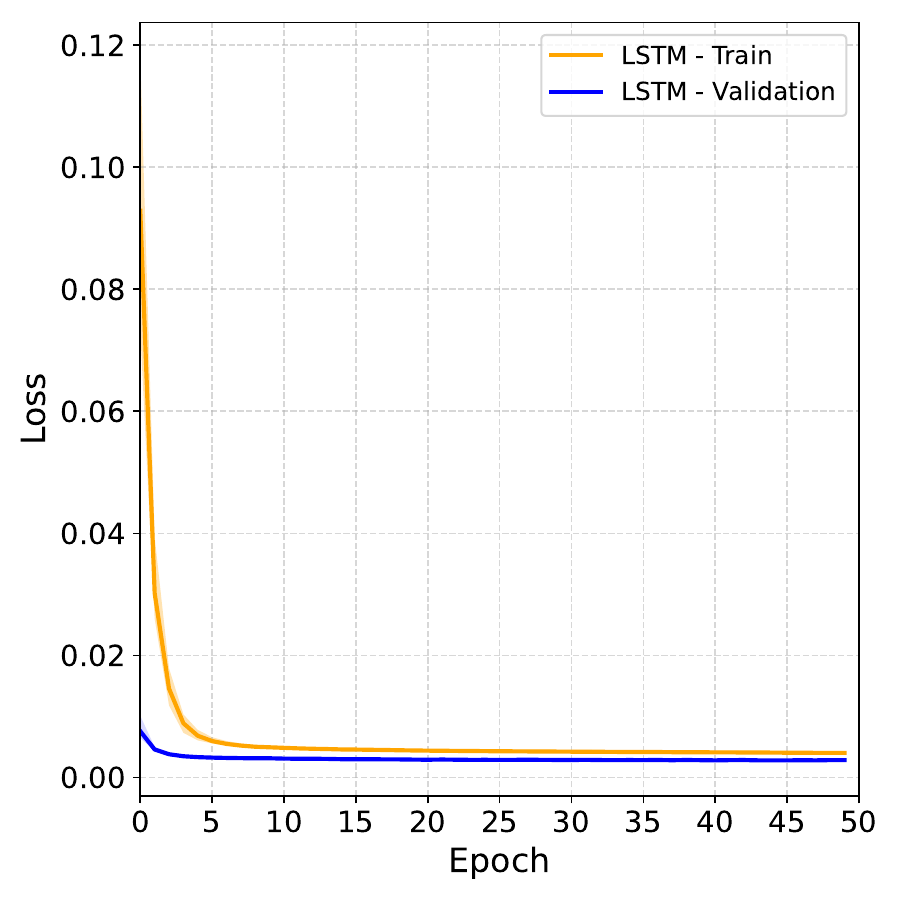}
        \caption{LSTM}
        \label{fig:exp-reg-lstm-curve}
    \end{subfigure}

    \hfill

    \begin{subfigure}[b]{0.45\textwidth}
        \centering
        \includegraphics[width=\textwidth]{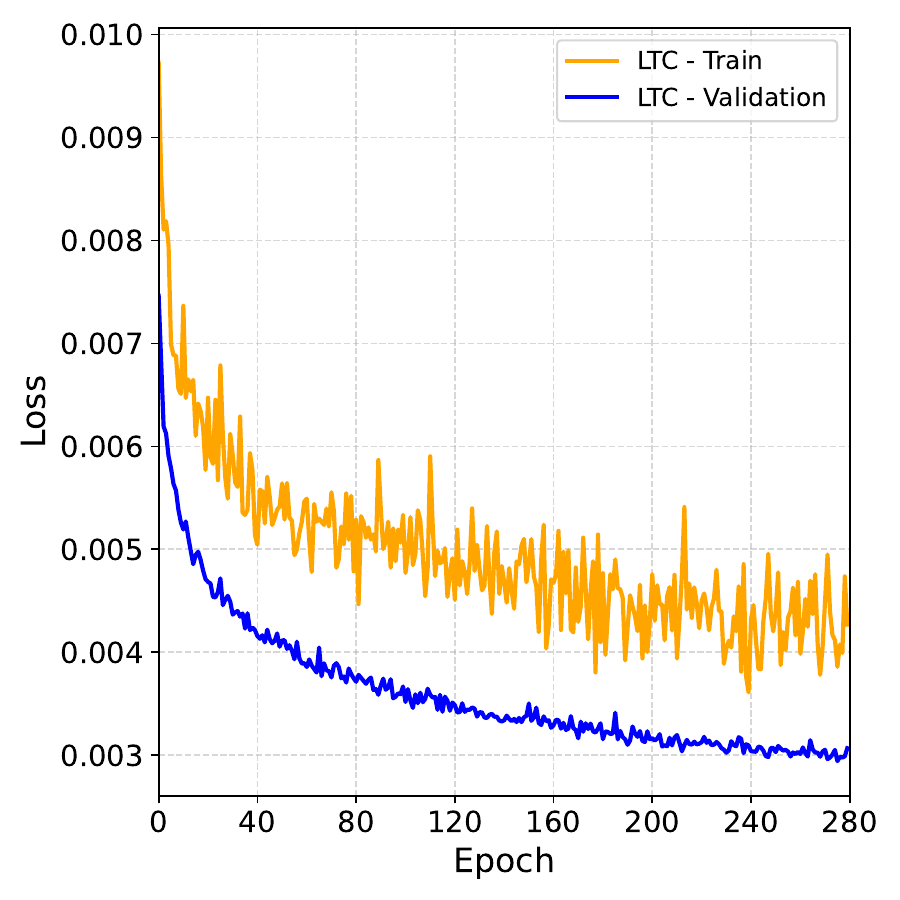}
        \caption{LTC}
        \label{fig:exp-reg-ltc-curve}
    \end{subfigure}
    \hspace{5mm}
    \begin{subfigure}[b]{0.45\textwidth}
        \centering
        \includegraphics[width=\textwidth]{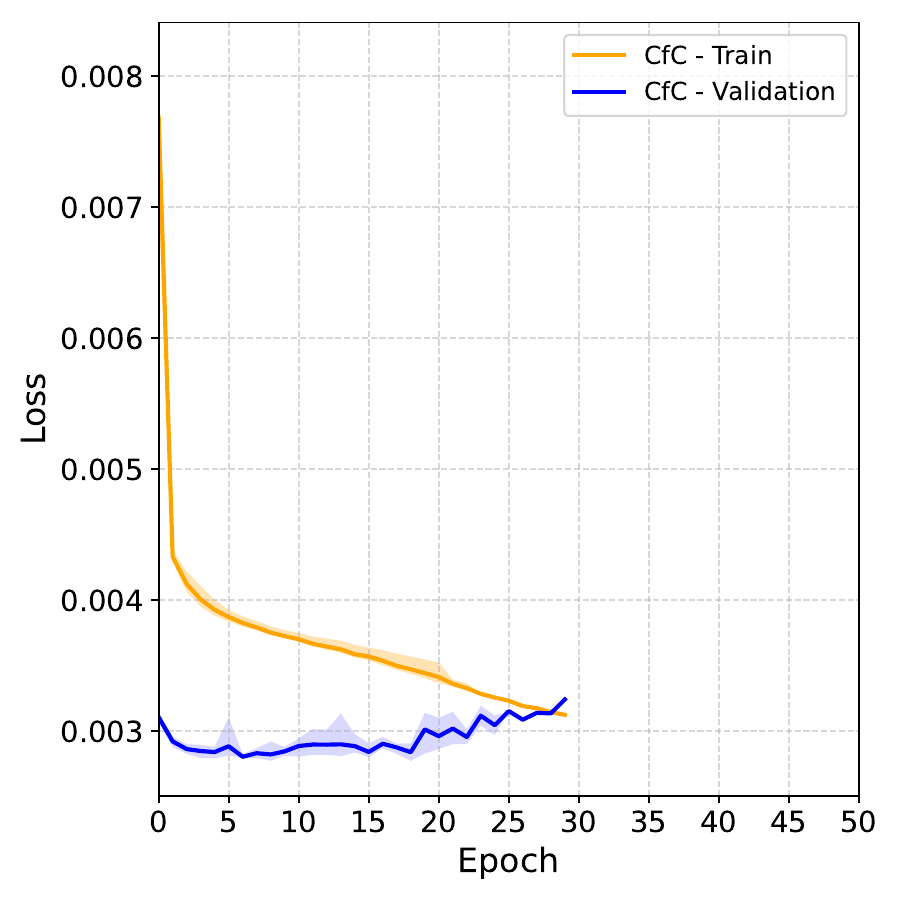}
        \caption{CfC}
        \label{fig:exp-reg-cfc-curve}
    \end{subfigure}

    \caption{Averaged loss curves obtained on \textit{rgc9} dataset in the \textit{base} regression task by all evaluated models.}
    \label{fig:exp-reg-loss-curves}
\end{figure}

In addition to achieving the best correlation, the \textit{ConvNet} model also showed the most stable learning process among all models considered. Figure~\ref{fig:exp-reg-loss-curves} shows the averaged training and validation losses for each model obtained on the \textit{rgc9} dataset. For the \textit{ConvNet} model, loss curves for both sets generally align with each other, smoothly decreasing with each successive epoch. In contrast, the LTC and CfC models continuously minimized the loss on the training set, while from the beginning of training, they obtained a low loss on the validation set, which they were no longer able to minimize successively during the training process. Such behavior suggests that these models quickly overfit the data, and most of the actual learning occurs in the first training epoch.

Apart from monitoring the MSE loss during training, the correlation on the validation set was measured every epoch. Figure~\ref{fig:exp-reg-valid-cor} shows the averaged correlation achieved by each model for each dataset in successive epochs of the training. Both the \textit{ConvNet} model and the Neural ODE-based models quickly achieved a relatively high correlation on the validation set, given the nature of the undertaken task. The exception here was the LSTM model, which needed significantly more time to achieve a promising level of correlation.

The subpar performance of LTC and CfC might be attributed to their continuous nature and the almost binary character of the target variable. Although the retina is a dynamical system, the fact that the firing rates are available only in the binned form prevents taking advantage of such features of the NODE-based architectures as sampling data at irregular time points.

In brief, this series of experiments shows that a simple model outperforms more refined recurrent architectures. This may suggest that the time dimension is not as crucial in this task, or that the temporal information has not been fully exploited by RNNs in the presented setup. Arguably, mapping the information from 40 video frames through a convolutional stack and a dense layer to a 32-dimensional vector does not provide an opportunity to leverage the sequential processing power of RNNs. Hence, we enable sequential processing in another experiment outlined in Sec.~\ref{sec:exp-reg-multiscale-temp}.

\begin{table}[t!]
\centering
\caption{Results obtained by the CfC model trained in four different configurations of multi-scale temporal representation. The notation used together with the model name indicates the type of the configuration, where the first number in the subscript denotes $M$ and the second $N$ parameters. For example, the configuration CfC\textsubscript{$2x20$} means that the CfC model was trained with a sequence of size $2$, where each element in a sequence contained $20$ consecutive historical visual frames.}
\label{tab:cfc-multiscale}
\begin{tabular}{
    ll
    S[table-format=1.2(2),
    separate-uncertainty = true,
    table-number-alignment = right,
	table-figures-uncertainty = 1,]
    S[table-format=2.2(2),
    separate-uncertainty = true,
	table-figures-uncertainty = 1,
    table-number-alignment = right]
    S[table-format=6.0,
    table-number-alignment = right]
    S[table-format=4.0(1),
    group-digits = integer,
    group-separator = {,},
    group-minimum-digits = 3,
    separate-uncertainty = true,
	table-figures-uncertainty = 1,
    table-number-alignment = center,
    table-column-width=2cm]
}
\toprule
\textbf{Dataset} & \textbf{Model} & \textbf{$\rho$ \textuparrow} & \textbf{MAE \textdownarrow} & \textbf{\# of params} & \textbf{Time [s]} \\
\midrule
\multirow{4}{*}{\textit{rgc9}}
  & \textbf{CfC\textsubscript{$1x40$}} & \bfseries 0.47(1) & \bfseries 2.86(7) & \bfseries 47496 & \bfseries 2411(387) \\
  & CfC\textsubscript{$2x20$} & 0.33(6) & 2.53(5) & 44616 & 1433(240) \\
  & CfC\textsubscript{$4x10$} & 0.32(4) & 2.53(12) & 43176 & 2270(382) \\
  & CfC\textsubscript{$8x5$} & 0.18(10) & 2.47(7) & 42888 & 3256(1269) \\
\midrule
\multirow{4}{*}{\textit{rgc14}}
  & \textbf{CfC\textsubscript{$1x40$}} & \bfseries 0.55(1) & \bfseries 2.35(6) & \bfseries 50752 & \bfseries 2673(300) \\
  & CfC\textsubscript{$2x20$} & 0.36(13) & 2.50(11) & 47872 & 2736(1565) \\
  & CfC\textsubscript{$4x10$} & 0.28(2) & 2.36(5) & 46432 & 3029(1111) \\
  & CfC\textsubscript{$8x5$} & 0.24(12) & 2.30(11) & 46144 & 2830(412) \\
\midrule
\multirow{4}{*}{\textit{rgc27}}
  & \textbf{CfC\textsubscript{$1x40$}} & \bfseries 0.52(1) & \bfseries 3.03(13) & \bfseries 55144 & \bfseries 3259(610) \\
  & CfC\textsubscript{$2x20$} & 0.37(4) & 2.93(14) & 52264 & 3098(821) \\
  & CfC\textsubscript{$4x10$} & 0.27(3) & 3.02(7) & 50824 & 3814(762) \\
  & CfC\textsubscript{$8x5$} & 0.22(5) & 3.07(9) & 50536 & 4945(1127) \\
\bottomrule
\end{tabular}
\end{table}

\begin{figure}[t]
    \centering

    \begin{subfigure}[b]{0.45\textwidth}
        \centering
        \includegraphics[width=\textwidth]{figures/ltc_loss_curve.pdf}
        \caption{Loss curve}
        \label{fig:appendix-ltc-small-lr-loss}
    \end{subfigure}
    \hspace{5mm}
    \begin{subfigure}[b]{0.45\textwidth}
        \centering
        \includegraphics[width=\textwidth]{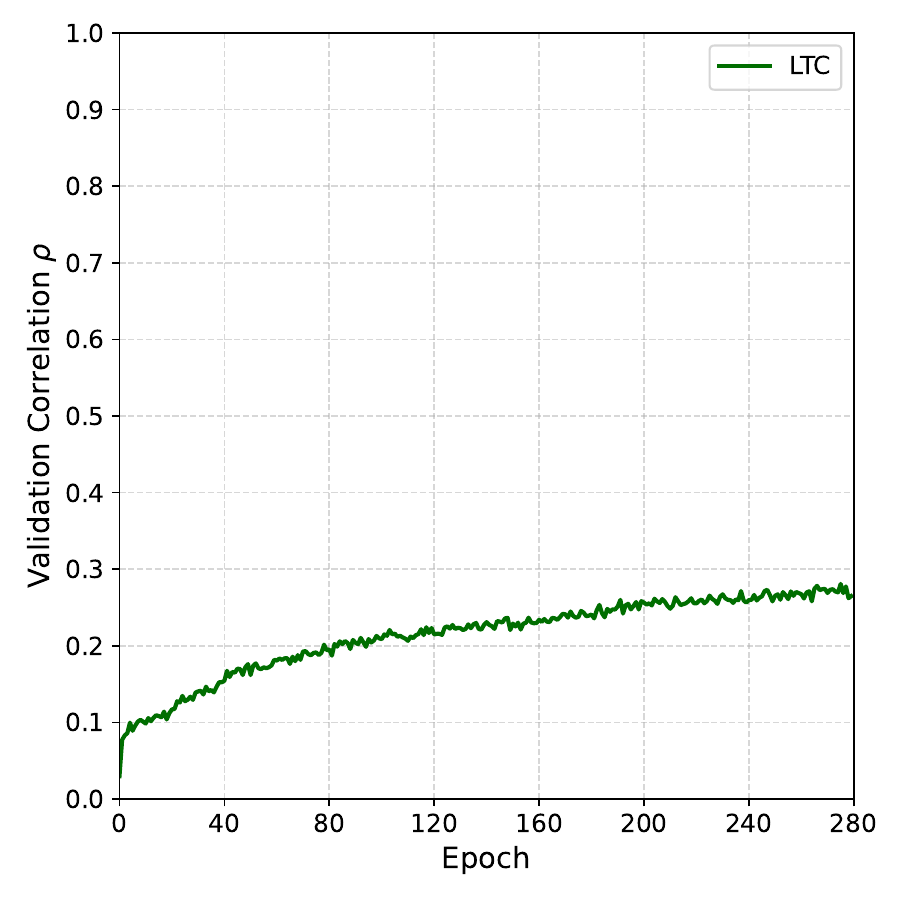}
        \caption{Correlation curve}
        \label{fig:appendix-ltc-small-lr-correlation}
    \end{subfigure}

    \caption{Loss function and correlation curves obtained during the LTC model training with a substantially smaller learning rate. The model was trained for $280$ epochs, which in this configuration stands for a single iteration over the training set. The model was validated every $1{,}024$ data samples.}
    \label{fig:appendix-ltc-small-lr}
\end{figure}
\begin{figure}[t]
    \centering
    \includegraphics[width=.9\linewidth]{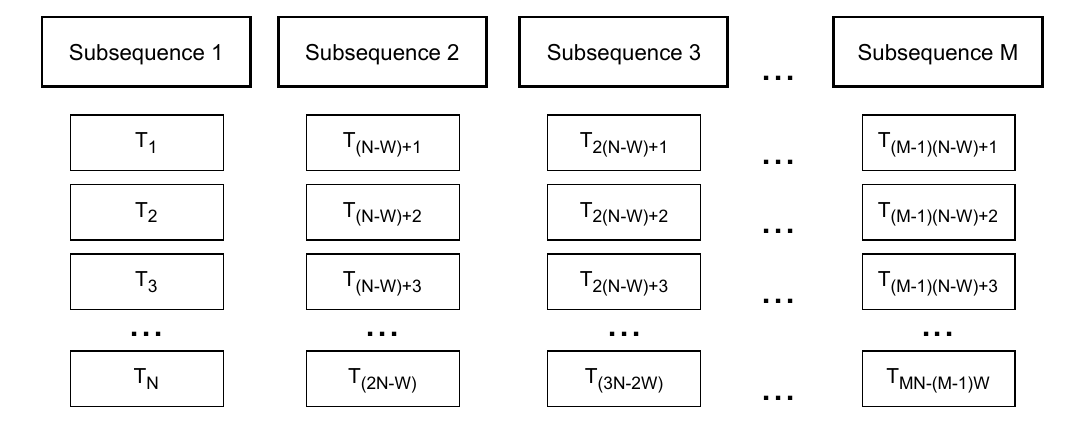}
    \caption{A diagram explaining the concept behind two-scale temporal representation in RNN-based architectures. The recurrent block processes $M$ subsequences, where each subsequence contains $N$ succeeding grayscale images stacked along the channel dimension in a single input image. A $W$ parameter specifies the number of frames shared between two adjacent subsequences. In total, the neural network operates on $M \times N - (M-1) \times W$ visual frames, which simplifies to $M \times N$ for the case with $W=0$.}
    \label{fig:two-scale-time-schema}
\end{figure}

\section{Investigation of training efficiency in LTCs} \label{sec:reg-ltc-small-lr}

What stands out in Fig.~\ref{fig:exp-reg-loss-curves} is that most of the learning process in NODEs occurs in the first training epoch. To better understand the behavior of these architectures, we performed an additional experiment with the LTC network. In this experiment, the learning rate of the LTC block was decreased from the default value of $2\times10^{-3}$ to $1\times10^{-6}$. Moreover, the model was validated every $1{,}024$ training samples instead of every full iteration over the training set.
The loss and correlation curves are shown in Figs.\ \ref{fig:appendix-ltc-small-lr-loss} and \ref{fig:appendix-ltc-small-lr-correlation}, respectively. Although the loss function on the validation and training sets was much more streamlined compared to the \textit{base} configurations presented in the previous section (Sec.\ \ref{sec:reg-base-results}), finally, the model reached a worse Pearson correlation equal to $0.41$. Nevertheless, a surprising observation is that even though a single epoch contained only $1{,}024$ data samples, the model converged very quickly to a small validation loss. Remarkably, the model learned most of the dynamics in the first $20$ epochs, which corresponds to merely about $7\%$ of the entire training set. On the one hand, this may suggest that NODEs need much smaller data sets to generalize well, but on the other hand, it may mean that they fall quite quickly into a local suboptimum from which they are unable to escape in the further stages of training.

\section{Multi-scale temporal representation} \label{sec:exp-reg-multiscale-temp}

We devised an extended version of NODE-based architectures that used two time scales: coarse and fine. At the coarse scale, a sequence of time steps is processed by the recurrent block of the neural network. On the fine scale, a stack of consecutive visual frames is combined into a single image. To distinguish these two scales, the following nomenclature was introduced:

\begin{itemize}
    \item  \textbf{subsequence} -- a collection of $N$ consecutive grayscale images combined along the channel dimension into a single image (the fine scale),
    \item \textbf{sequence} -- an ordered set of $M$ subsequences processed over time by the RNN (the coarse scale),
    \item \textbf{subsequence overlap} -- the number of $W$ consecutive frames shared between two adjacent subsequences within a sequence, $W \in \{0,1,...,N-1\}$.
\end{itemize}
In the basic configuration, $W$ was set to $0$, meaning no overlapping frames between subsequences. Hence, the neural network operated on $M \times N$ visual frames by default. The graphical form of the introduced concept is presented in Fig.~\ref{fig:two-scale-time-schema}.

In order to ensure a fair comparison, the total number of processed images in the two-scale temporal representation scenario was kept equal to that of the default configuration, where a single input contained consecutive stacked images. Therefore, the original value of $N=40$ was divided such that $M \times N = 40$, leading to distinguishing the following configurations:
\begin{itemize}
    \item $M=2, N=20$,
    \item $M=4, N=10$,
    \item $M=8, N=5$.
\end{itemize}
In these experiments, only the Neural ODE-based architectures were considered as more sophisticated representatives of RNNs. Both the LTC and CfC models were trained in the same configuration as on the \textit{base} regression task. For brevity, Table \ref{tab:cfc-multiscale} shows the results obtained for the CfC model only.

Remarkably, regardless of the dataset, a counterintuitive tendency can be observed in the results -- increasing the sequence length deteriorates the predictive capabilities. In other words, the model using the sequence of one performs substantially better than a model processing the same input frames distributed into eight sequences of five images. Our results cast new light on the nature of temporal dependencies in the dataset. Specifically, such results may suggest that RNN architectures, which naturally learn temporal dependencies over time, are not as accurate as models that focus more on the spatiotemporal integration and optimize filters that combine temporal and spatial information jointly using the convolutional layers.

\section{Hyperparameter optimization} \label{sec:hpo-exp}

\begin{table}[t!]
\centering
\caption{Parameters of the HPO experiment. Learning rates are sampled from a uniform distribution.}
\begin{tabular}{ll}
\toprule
\textbf{Parameter } & \textbf{Values/Distribution} \\
\midrule
Subsequence length & 20, 30, 40 \\
Batch size & 2048, 4096, 8192 \\
Latent size & 16, 32, 64 \\
Encoder lr & Uniform (1e-5, 0.1) \\
RNN hidden size & 12, 14, 16, 20 \\
Predictor lr & Uniform (0.0001, 0.3) \\
RNN type & \texttt{LTC}, \texttt{CfC} \\
\bottomrule
\end{tabular}
\label{tab:hpo-rnn}
\end{table}

For hyperparameter optimization (HPO), we opted for the Weights \& Biases Sweeps feature \cite{wandb}. A key advantage of this approach is the flexibility to select the performance metric from the metrics logged to the WandB service, in our case, the Pearson correlation on the validation set. Common HPO strategies include random search, grid search, and Bayesian optimization. Bayesian optimization, which is an informed search method that determines the set of parameters to evaluate in subsequent iterations based on a surrogate function, was chosen for both experiments \cite{hpo-hyperband}. We confined our optimization efforts to the \textit{rgc9} dataset because of the computational demands involved. Nonetheless, given the similarity to the other two datasets, it is probable that the outcomes will be applicable to them as well.

\begin{figure}[t]
    \centering
    \includegraphics[width=\linewidth]{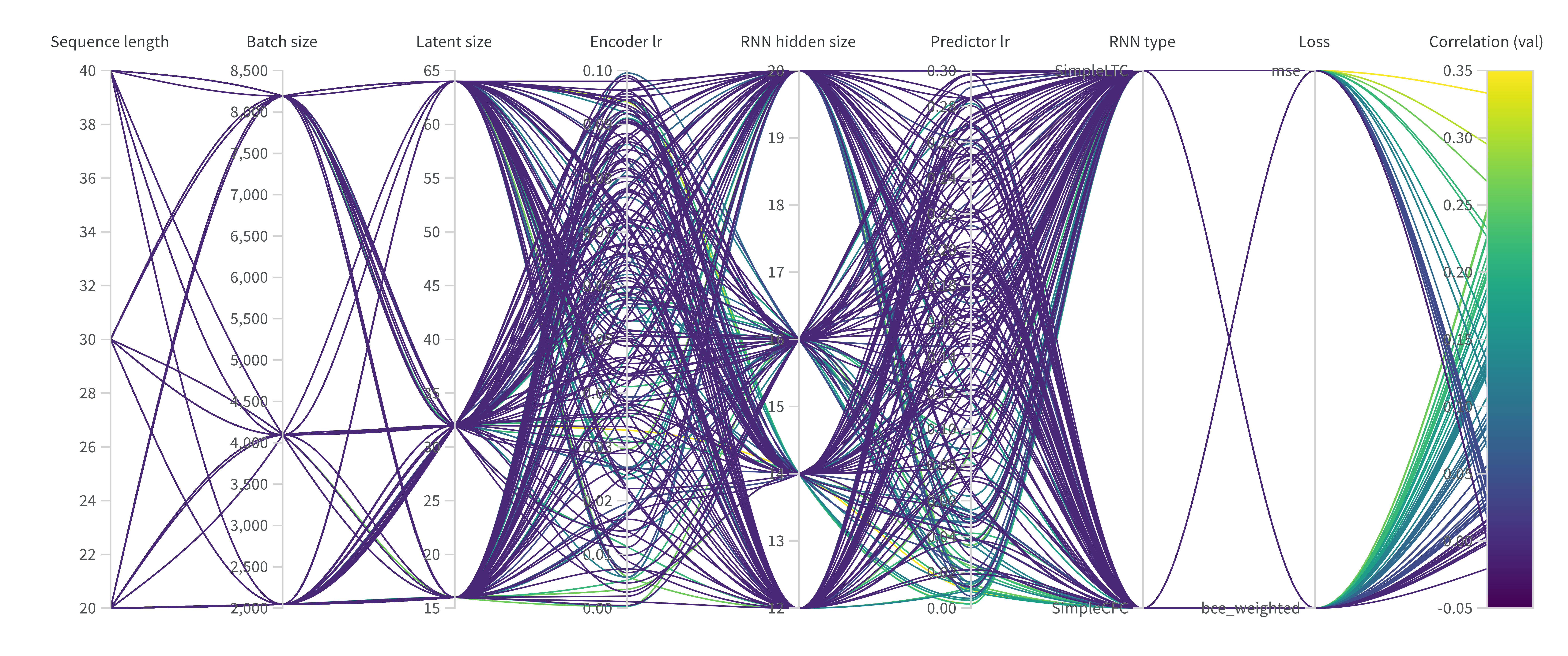}
    \caption{Configurations explored over 24 hours of the HPO with the final validation set correlation (right).}
    \label{fig:hpo-basic}
\end{figure}

This series of experiments explored a vast space of parameters to possibly create the strongest NODE-based model. For a complete list of tuned hyperparameters, please refer to Table~\ref{tab:hpo-rnn}. The stopping condition for the search was 24 hours. Throughout this time, 197 runs with various combinations of parameters were subject to training and evaluation, as presented in Fig.\ \ref{fig:hpo-basic}. The best configuration attained $\rho=0.33$ on the validation set and represents a setup akin to the default one (presented in Sec.~\ref{sec:exp-node-models}). The complete list of hyperparameters selected by HPO can be seen below:

\begin{itemize}[noitemsep]
    \item subsequence length = 40,
    \item batch size = $2,048$ (default was $4,096$),
    \item latent size = $32$,
    \item encoder learning rate = $0.033$ (default was $0.001$),
    \item predictor learning rate = $0.01$ (default was $0.002$),
    \item predictor hidden size = 14.
\end{itemize}

\begin{figure}[t]
    \centering
    \includegraphics[width=0.5\linewidth]{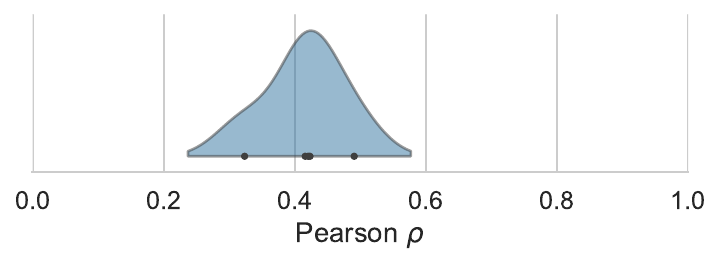}
    \caption{Test-set correlation of five models trained with the best hyperparameters discovered through HPO.}
    \label{fig:exp-hpo-violin-basic}
\end{figure}

After conducting the extensive HPO experiment, five models with the optimal parameters were trained. The evaluation on the testing set showed $\rho=0.41 \ (\pm 0.06)$, with the top-performing model achieving a score of 0.49 and exceeding the mean score of CfC model as seen in Table \ref{tab:regression-results} of the main text. Figure~\ref{fig:exp-hpo-violin-basic} presents the distribution of $\rho$ values for the optimized models.

On balance, the hyperparameter optimization experiment offers interesting insights into the parametrization of the presented models leading to, for example, low variance on the testing set. While these outcomes do not surpass the basic models discussed in Sec. \ref{sec:reg-base-results}, this examination provides us with more intuition on the orchestration of the encoder and RNN components.

\section{Noise robustness analysis} \label{sec:reg:noise-robus-exp}

\begin{figure}[t]
    \centering
    \includegraphics[width=0.8\linewidth]{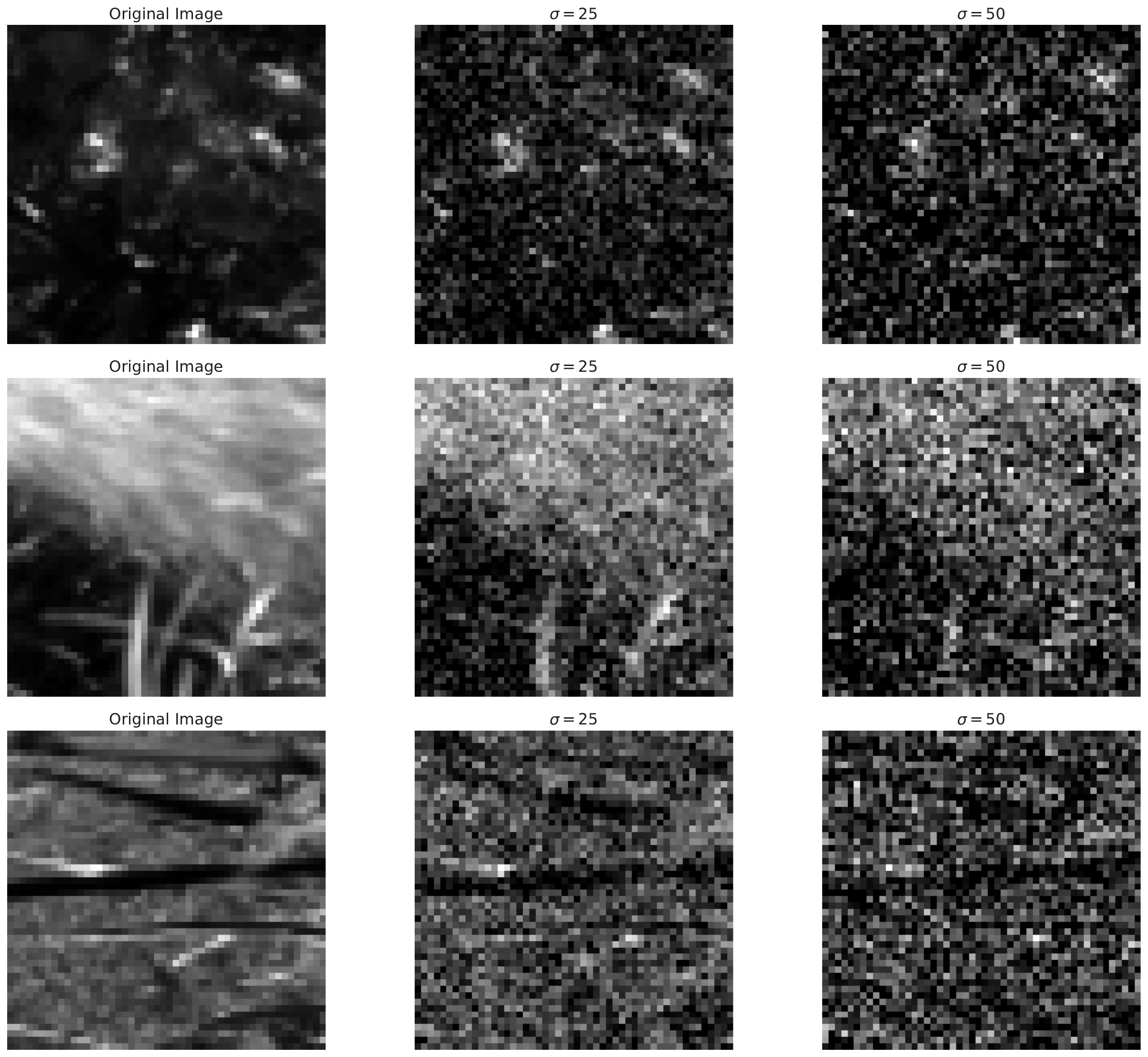}
    \caption{Sample original (non-perturbed) testing frames and their noisy versions.}
    \label{fig:exp-noised-images}
\end{figure}

Following Lechner et al.\ \cite{theory/ncp}, we conduct a robustness evaluation. The cited work showcased the noise robustness of the NCP in preventing crashes within an autonomous driving context. Consequently, we investigate whether our NCP-based models, the LTC and CfC, maintain their performance when subjected to substantial input perturbations, specifically the application of pixel-wise Gaussian noise with zero mean. Using a methodology adapted from the cited study, we assess model performance through the relative difference in correlation scores on the test set between non-perturbed and perturbed inputs. Our analysis involves two noise scenarios: moderate ($\sigma=25$) and severe ($\sigma=50$), where any values exceeding 255 are clipped. Figure \ref{fig:exp-noised-images} illustrates example non-perturbed images along with their noisy counterparts.

\begin{table}[t!]
\centering
\caption{Noise robustness analysis on the \emph{rgc9} data set, where the input frames were noised with Gaussian noise ($\mu = 0$) with $\sigma = 25$ and $\sigma = 50$. The best models across five runs were selected for comparison and evaluated against the noisy testing set. Model quality is expressed in the mean Pearson $\rho$ coefficient over all output channels. The row in bold corresponds to the model with the lowest relative difference between the correlation for the non-perturbed data and the noisy data.}
\label{tab:noise-robustness}
\begin{tabular}{@{}lrrrrr@{}}
\toprule
Model & Non-perturbed & $\sigma = 25$ & Relative difference {[}\%{]} & $\sigma = 50$ & Relative difference {[}\%{]} \\ \midrule
\textbf{ConvNet} & \textbf{0.57} & \textbf{0.55} & \textbf{-3.11} & \textbf{0.51} & \textbf{-10.94} \\
LSTM & 0.45 & 0.43 & -3.94 & 0.40 & -11.24 \\
LTC & 0.49 & 0.47 & -5.58 & 0.42 & -14.88 \\
CfC & 0.47 & 0.45 & -4.44 & 0.40 & -15.87 \\ \bottomrule
\end{tabular}
\end{table}

We carried out the analysis by testing the top models chosen from five iterations in the standard configuration (subsequence length $= 40$) using the original, non-perturbed \textit{rgc9} dataset. More precisely, we used the basic models described in Sec.~\ref{sec:reg-base-results}, and the best instances were selected from the models discussed in Sec. \ref{sec:reg-base-results}. In contrast to the conclusions drawn by Lechner et al. and our initial expectations, NCP models do not exhibit severe noise resiliency, as demonstrated in Table~\ref{tab:noise-robustness}. Remarkably, ConvNet and LSTM demonstrate a more favorable relative difference. This might indicate that LTCs and CfCs do not adequately capture the underlying data patterns during training, limiting their ability to generalize in more complex situations. Alternatively, it could be that the noise robustness observed in Lechner's experiments does not extend to all scenarios, particularly the complex task of modeling ganglion cells.

\section{Inference time} \label{sec:exp-inference-time}

We conducted measurements of inference time on NVIDIA A100 (with GPU synchronization to verify that all tasks are finalized before pausing the timer). The experiment was performed for the same model instances as in Sec.~\ref{sec:reg:noise-robus-exp} on the training and testing sets, with the results shown in Table~\ref{tab:inference-time}. This analysis found evidence for remarkably faster inference of LTC and CfC in comparison to \textit{ConvNet} and LSTM, with a larger gap exhibited on a larger dataset. Overall, this is a promising finding for the application of NODE-based architectures in retinal prostheses and other fields, in which a real-time mode of operation is crucial.

\begin{table}[t!]
\centering
\caption{Average inference time per data point in seconds. We performed time measurements on both the training and testing data, which differ only in size, equal to $287,801$ and $5,956$, respectively. The speed advantage of the NODE-based models is more prominent on the bigger dataset.
}
\label{tab:inference-time}
\begin{tabular}{@{}lrr@{}}
\toprule
Model & Testing data time [s] & Training data time [s] \\ \midrule
ConvNet & \num{2.34e-4} & \num{23e-6} \\
LSTM &  \num{2.60e-4} & \num{23e-6} \\
\textbf{LTC} & \textbf{\num{2.29e-4}} & \textbf{\num{5e-6}} \\
\textbf{CfC} & \textbf{\num{2.30e-4}} & \textbf{\num{5e-6}} \\ \bottomrule
\end{tabular}
\end{table}

\bibliographystyle{unsrtnat}
\bibliography{references}  

\begin{thebibliography}{17}
\providecommand{\natexlab}[1]{#1}
\providecommand{\url}[1]{\texttt{#1}}
\expandafter\ifx\csname urlstyle\endcsname\relax
  \providecommand{\doi}[1]{doi: #1}\else
  \providecommand{\doi}{doi: \begingroup \urlstyle{rm}\Url}\fi

\bibitem[Chen et~al.(2019)Chen, Rubanova, Bettencourt, and Duvenaud]{theory/chen2019neuralordinarydifferentialequations}
Ricky T.~Q. Chen, Yulia Rubanova, Jesse Bettencourt, and David Duvenaud.
\newblock Neural ordinary differential equations, 2019.
\newblock URL \url{https://arxiv.org/abs/1806.07366}.

\bibitem[Hasani et~al.(2021)Hasani, Lechner, Amini, Rus, and Grosu]{theory/Hasani20217657}
Ramin Hasani, Mathias Lechner, Alexander Amini, Daniela Rus, and Radu Grosu.
\newblock Liquid time-constant networks.
\newblock In \emph{35th AAAI Conference on Artificial Intelligence, AAAI 2021}, volume~9A, page 7657 – 7666, 2021.
\newblock URL \url{https://www.scopus.com/inward/record.uri?eid=2-s2.0-85109163277&partnerID=40&md5=8fa22558aacae5c43a204ba76f10f9c0}.
\newblock Cited by: 84.

\bibitem[Hasani et~al.(2022)Hasani, Lechner, Amini, Liebenwein, Ray, Tschaikowski, Teschl, and Rus]{theory/Hasani2022992}
Ramin Hasani, Mathias Lechner, Alexander Amini, Lucas Liebenwein, Aaron Ray, Max Tschaikowski, Gerald Teschl, and Daniela Rus.
\newblock Closed-form continuous-time neural networks.
\newblock \emph{Nature Machine Intelligence}, 4\penalty0 (11):\penalty0 992 – 1003, 2022.
\newblock \doi{10.1038/s42256-022-00556-7}.
\newblock URL \url{https://www.scopus.com/inward/record.uri?eid=2-s2.0-85141901775&doi=10.1038%2fs42256-022-00556-7&partnerID=40&md5=3be25fa8c86fba70777b304c3715810c}.
\newblock Cited by: 34; All Open Access, Hybrid Gold Open Access.

\bibitem[Maheswaranathan et~al.(2023)Maheswaranathan, McIntosh, Tanaka, Grant, Kastner, Melander, Nayebi, Brezovec, Wang, Ganguli, et~al.]{retina/maheswaranathan2023interpreting}
Niru Maheswaranathan, Lane~T McIntosh, Hidenori Tanaka, Satchel Grant, David~B Kastner, Joshua~B Melander, Aran Nayebi, Luke~E Brezovec, Julia~H Wang, Surya Ganguli, et~al.
\newblock Interpreting the retinal neural code for natural scenes: From computations to neurons.
\newblock \emph{Neuron}, 111\penalty0 (17):\penalty0 2742--2755, 2023.

\bibitem[Hochreiter and Schmidhuber(1997)]{theory/Hochreiter19971735}
Sepp Hochreiter and Jürgen Schmidhuber.
\newblock Long short-term memory.
\newblock \emph{Neural Computation}, 9\penalty0 (8):\penalty0 1735 – 1780, 1997.
\newblock \doi{10.1162/neco.1997.9.8.1735}.
\newblock URL \url{https://www.scopus.com/inward/record.uri?eid=2-s2.0-0031573117&doi=10.1162%2fneco.1997.9.8.1735&partnerID=40&md5=6e4ee65c4bc5399487e5a65f4186aa19}.
\newblock Cited by: 74221.

\bibitem[Kingma and Ba(2017)]{training/adam}
Diederik~P. Kingma and Jimmy Ba.
\newblock Adam: A method for stochastic optimization, 2017.
\newblock URL \url{https://arxiv.org/abs/1412.6980}.

\bibitem[MATLAB(2010)]{problem/MATLAB:2010}
MATLAB.
\newblock \emph{version 7.10.0 (R2010a)}.
\newblock The MathWorks Inc., Natick, Massachusetts, 2010.

\bibitem[Brainard(1997)]{problem/ThePsychophysicsToolbox}
David~H. Brainard.
\newblock The psychophysics toolbox.
\newblock \emph{Spatial Vision}, 10\penalty0 (4):\penalty0 433 -- 436, 1997.
\newblock \doi{10.1163/156856897X00357}.
\newblock URL \url{https://brill.com/view/journals/sv/10/4/article-p433_15.xml}.

\bibitem[Tkačik et~al.(2011)Tkačik, Garrigan, Ratliff, Milčinski, Klein, Seyfarth, Sterling, Brainard, and Balasubramanian]{problem/natural-scene-images}
Gašper Tkačik, Patrick Garrigan, Charles Ratliff, Grega Milčinski, Jennifer~M. Klein, Lucia~H. Seyfarth, Peter Sterling, David~H. Brainard, and Vijay Balasubramanian.
\newblock Natural images from the birthplace of the human eye.
\newblock \emph{PLOS ONE}, 6\penalty0 (6):\penalty0 1--12, 06 2011.
\newblock \doi{10.1371/journal.pone.0020409}.
\newblock URL \url{https://doi.org/10.1371/journal.pone.0020409}.

\bibitem[Ribeiro(2009)]{Ribeiro2009}
Carla~Isabel Ribeiro.
\newblock Ambystoma tigrinum (tiger salamander), 2009.
\newblock URL \url{https://commons.wikimedia.org/wiki/File:Salamandra_Tigre.png}.
\newblock Own work, accessed 29 May 2025.

\bibitem[Glorot and Bengio(2010)]{training/xavier}
Xavier Glorot and Yoshua Bengio.
\newblock Understanding the difficulty of training deep feedforward neural networks.
\newblock In \emph{Proceedings of the thirteenth international conference on artificial intelligence and statistics}, pages 249--256. JMLR Workshop and Conference Proceedings, 2010.

\bibitem[Saxe et~al.(2014)Saxe, McClelland, and Ganguli]{training/orthogonal}
Andrew~M. Saxe, James~L. McClelland, and Surya Ganguli.
\newblock Exact solutions to the nonlinear dynamics of learning in deep linear neural networks, 2014.
\newblock URL \url{https://arxiv.org/abs/1312.6120}.

\bibitem[Paszke et~al.(2019)Paszke, Gross, Massa, Lerer, Bradbury, Chanan, Killeen, Lin, Gimelshein, Antiga, Desmaison, Kopf, Yang, DeVito, Raison, Tejani, Chilamkurthy, Steiner, Fang, Bai, and Chintala]{pytorch}
Adam Paszke, Sam Gross, Francisco Massa, Adam Lerer, James Bradbury, Gregory Chanan, Trevor Killeen, Zeming Lin, Natalia Gimelshein, Luca Antiga, Alban Desmaison, Andreas Kopf, Edward Yang, Zachary DeVito, Martin Raison, Alykhan Tejani, Sasank Chilamkurthy, Benoit Steiner, Lu~Fang, Junjie Bai, and Soumith Chintala.
\newblock Pytorch: An imperative style, high-performance deep learning library.
\newblock In \emph{Advances in Neural Information Processing Systems 32}, pages 8024--8035. Curran Associates, Inc., 2019.
\newblock URL \url{http://papers.neurips.cc/paper/9015-pytorch-an-imperative-style-high-performance-deep-learning-library.pdf}.

\bibitem[Biewald(2020)]{wandb}
Lukas Biewald.
\newblock Experiment tracking with weights and biases, 2020.
\newblock URL \url{https://www.wandb.com/}.
\newblock Software available from wandb.com.

\bibitem[Yadan(2019)]{Yadan2019Hydra}
Omry Yadan.
\newblock Hydra - a framework for elegantly configuring complex applications.
\newblock Github, 2019.
\newblock URL \url{https://github.com/facebookresearch/hydra}.

\bibitem[Lechner et~al.(2020)Lechner, Hasani, Amini, Henzinger, Rus, and Grosu]{theory/ncp}
Mathias Lechner, Ramin~M. Hasani, Alexander Amini, Thomas~A. Henzinger, Daniela Rus, and Radu Grosu.
\newblock Neural circuit policies enabling auditable autonomy.
\newblock \emph{Nat. Mach. Intell.}, 2\penalty0 (10):\penalty0 642--652, 2020.
\newblock \doi{10.1038/S42256-020-00237-3}.
\newblock URL \url{https://doi.org/10.1038/s42256-020-00237-3}.

\bibitem[Li et~al.(2018)Li, Jamieson, DeSalvo, Rostamizadeh, and Talwalkar]{hpo-hyperband}
Lisha Li, Kevin Jamieson, Giulia DeSalvo, Afshin Rostamizadeh, and Ameet Talwalkar.
\newblock Hyperband: A novel bandit-based approach to hyperparameter optimization.
\newblock \emph{Journal of Machine Learning Research}, 18\penalty0 (185):\penalty0 1--52, 2018.
\newblock URL \url{http://jmlr.org/papers/v18/16-558.html}.

\end{thebibliography}

\end{document}